# Text Semantics to Flexible Design: A Residential Layout Generation Method Based on Stable Diffusion Model

Zijin Qiu [a,b], Jiepeng Liu [a,b], Yi Xia [a,b,c], Hongtuo Qi [a,b,d*], Pengkun Liu [a,b,e*]

[a] *Key Laboratory of New Technology for Construction of Cities in Mountain Area (Chongqing University), Ministry of Education, Chongqing 400045, China*

[b] *School of Civil Engineering, Chongqing University, Chongqing 400045, China*

[c] *Hong Kong Center for Construction Robotics, The Hong Kong University of Science and Technology, Hong Kong*

[d] *School of Civil Engineering, Xinyu University, Jiangxi 338004, China*

[e] *Department of Civil and Environmental Engineering, Carnegie Mellon University, Pittsburgh, PA, 15213, United States*

\* Corresponding Author: hitqht@163.com, pengkunl@andrew.cmu.edu

## Abstracts

Flexibility in the AI-based residential layout design remains a significant challenge, as traditional methods like rule-based heuristics and graph-based generation often lack flexibility and require substantial design knowledge from users. To address these limitations, we propose a cross-modal design approach based on the Stable Diffusion model for generating flexible residential layouts. The method offers multiple input types for learning objectives, allowing users to specify both boundaries and layouts. It incorporates natural language as design constraints and introduces ControlNet to enable stable layout generation through two distinct pathways. We also present a scheme that encapsulates design expertise within a knowledge graph and translates it into natural language, providing an interpretable representation of design knowledge. This comprehensibility and diversity of input options enable professionals and non-professionals to directly express design requirements, enhancing flexibility and controllability. Finally, experiments verify the flexibility of the proposed methods under multimodal constraints better than state-of-the-art models, even when specific semantic information about room areas or connections is incomplete.

## 1. Introduction and Background

Architectural design in the Architecture, Engineering, and Construction (AEC) industry, particularly for residential buildings, must address complex requirements such as structural integrity, construction methods, and energy efficiency. The conceptual design phase is especially challenging, requiring creativity, experience, and careful management of the topological and geometric constraints. Traditionally, architectural design relies on clients expressing their requirements, which architects then manually transform into design solutions based on experience, a process that is often cumbersome and time-consuming. This inefficiency arises because clients typically lack design expertise, leading to vague requirements expressed in natural language, which are insufficient to support effective design. Consequently, this study aims to explore how natural language can be used effectively in design by providing clients with structured templates to express their requirements while assisting architects in generating rule-based designs in response. This could be benefit for common people while designing their personal house. To address this purpose, the task must fulfill two key functions: firstly, it must ensure flexibility in input conditions, as clients who are not experts in architectural design—such as homeowners or individuals planning their personal residences—can only provide simple requirements, such as the number of rooms, sizes, and spatial relationships. Secondly, the generated output must be stable, diverse, and controllable. Controllability ensures that output quality is high and meets personalized user needs, while diversity provides a broad range of floor plans from which users can choose.

Computer-aided generative design methods offer a way to accelerate the design process. These can be divided into rule-based heuristic search methods and learning-based methods. Rule-based heuristics, such as evolutionary and growth algorithms [1,2], effectively solve straightforward problems but struggle with complex layout designs and implicit design rules that defy explicit parameterization. Moreover, their generation speed is constrained by the manual input of parameters, which is time-consuming. With the development of intelligent algorithms and AI-generated content, learning-based approaches, including deep learning techniques such as generative adversarial networks (GANs) and graph neural networks (GNNs) [3–7], have proven adept at managing complex architectural requirements and implicit design knowledge.

However, issues of flexibility during the design stage remain unresolved. Owing to limitations in visual model architectures, existing methods often severely restrict the user's ability to modify conditions. As illustrated in Fig. 1, the design input forms of common methods are divided into learning objectives and design constraints. These methods often limit the types of learning objectives to boundaries or layouts. Relying on a single learning objective restricts flexibility; users can only select from pre-designed apartment types, providing a preferred layout without the ability to define necessary boundaries. Designers, however, can abstract rough boundaries as design input. Therefore, it is essential to develop a method that accepts both layouts and boundaries as input. Moreover, existing studies often use rigid ways of expressing design constraints, which require users to have substantial design knowledge. Common representations, such as knowledge graphs, require users to abstract mathematical and connection constraints into graphical structures—a significant challenge for non-professional or inexperienced designers. They may find it difficult to specify detailed space allocations and may only provide general preferences or constraints, such as the number of rooms and partial connection relationships. Since these preferences are described in natural language, users cannot abstract design knowledge into formal representations. Designers are then required to supplement these requirements in detail to form complete design specifications. Therefore, it is necessary to explore how to use natural language to express design knowledge and provide users and designers with a way to generate constraints using natural language. This study aims to improve the flexibility of generative design and reduce the requirements for professionalism while maintaining controllability.

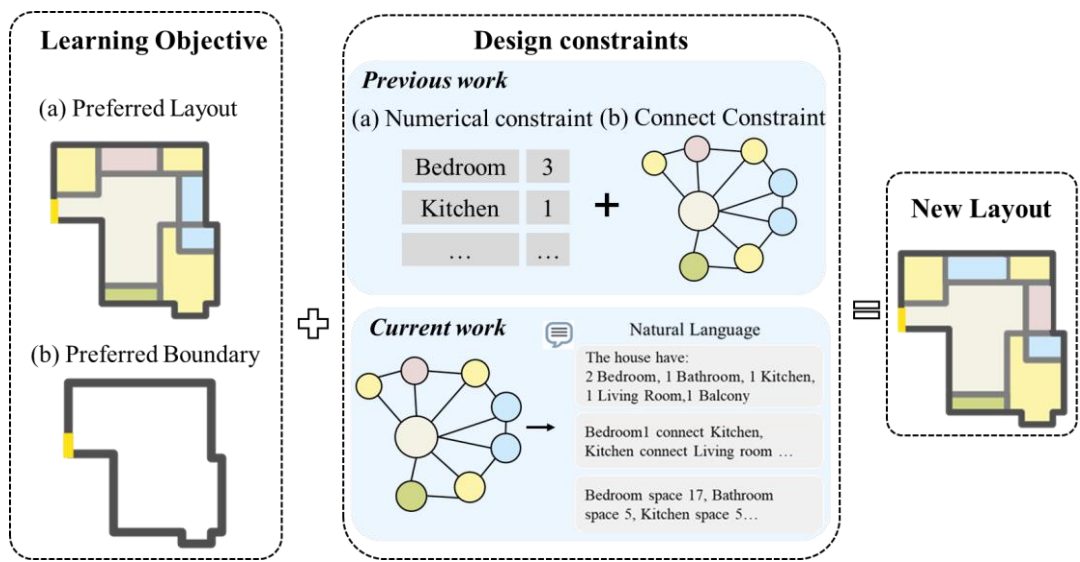

Fig.1. Input forms for generating residential layouts

To automate the design process for both professional and non-professional designers while maintaining flexibility, this paper proposes a cross-modal design approach based on the Stable Diffusion (SD) model. This method offers multiple input types for learning objectives, allowing users to specify boundaries or layouts. Regarding design constraints, we innovatively utilize natural language text and establish a paradigm for extracting design knowledge from knowledge graphs using natural language processing (NLP) techniques. This approach encapsulates design expertise within a knowledge graph, which is then translated into natural language, facilitating the standardized representation of design rules based on natural language. Our method reduces the need for extensive designer expertise and contrasts with traditional methods that require manual specification of knowledge graphs and double abstraction of design rules. Our approach intelligently combines two types of design constraints—natural language text and input images—to enhance the interpretability and flexibility of the design process, making it more accessible and controllable for all designers.

The study sets two specific objectives: (1) to define the design requirements, constraints, and evaluation metrics and to represent the design rules for residential layout design from design drawings in natural language and (2) to develop and evaluate a SD-based design method, assessing its effectiveness in generating innovative design solutions for residential layouts. In summary, the contributions of this research are manifold: (1) It introduces a methodology that translates complex architectural design rules into natural language, leveraging the synergy between knowledge graphs and NLP to enhance both the communicability and applicability of these rules in architectural projects; (2) It presents a cross-modal design methodology that utilizes Low-Rank Adaptation (LoRA) to fine-tune large-scale generative models, thereby enhancing their capacity to produce high-quality, compliant residential layouts from natural language inputs; (3) It demonstrates the generative model's adaptability and innovation in creating varied design solutions under conditions of incomplete semantic information, showcasing the model's ability to adjust creatively to missing data while adhering to design constraints. This flexibility is essential for practical applications where initial design information may be incomplete or subject to change.

The paper has the following structure: Section 2 discusses the related work; Section 3 defines the problem, relevant design rules, and evaluation metrics for architectural layout design; Section 4 explains the method for the SD-based design method, the dataset preparation; Section 5 reports a validation study that assesses the generated layouts; and Section 6 ends with final conclusions and suggestions for future research.

## 2. Literature Review

### 2.1 Architectural Layout Design Knowledge

As urbanization and populations rise, residential design has become increasingly important, blending a mix of taste, quality, and practicality while satisfying various stakeholders. Architectural layout design optimizes the arrangement of spaces to fulfill both function and form, balancing considerations like space use, flow, natural light, air quality, privacy, ease of access, and affordability. Residential layouts prioritize bedroom, bathroom, and living room configurations to match the number of potential residents and market trends, while adhering to regulations and what buyers want.

Architectural layout design knowledge refers to the information and rules that guide the creation and evaluation of spatial configurations for buildings. It includes aspects such as the functional requirements, spatial relations, circulation patterns, environmental factors, aesthetic preferences, and cultural influences of a design problem. Architectural layout design knowledge can be derived from various sources, such as precedents, standards, guidelines, codes, and user feedback. It can also be acquired through the experience, intuition, and creativity of the designer. For example, Tokunaga and Murota [8] noted that consumer preferences regarding unit size and type are critical determinants of a building's appeal, with varying preferences observed across different life stages of residents. The interconnectivity of spaces significantly influences the quality of residential design. Well-connected spaces enhance community living and reduce the occurrence of isolated areas, which is critical for promoting resident mobility and comfort. Weber et al. [9] emphasize the benefits of connected spaces, highlighting the importance of efficient circulation paths and minimizing empty or unusable areas.

Different sources of knowledge, such as design guidelines, precedents, standards, and user feedback, can inform and inspire the generation of design alternatives. However, managing and applying this knowledge effectively is not trivial, as it requires a systematic and structured approach to represent and evaluate design knowledge.

## 2.2 Knowledge Representation from Designs

One of automated design's main difficulties is capturing and expressing domain-specific knowledge and constraints, such as those in residential architecture. Knowledge representation is the method of defining and structuring the data and rules of a problem domain in a machine-readable way [10]. It allows computers to think about the design problem and create solutions that meet the given needs and goals.

One of the common ways to represent design rules is through shape grammars, which are formal systems that define the rules and operations for manipulating shapes [11]. Shape grammars can capture the syntactic and semantic aspects of design, enabling the generation of layouts that conform to a specific style or typology. Another method uses optimization techniques, which are mathematical methods for finding the best design solution among a set of feasible alternatives according to an objective function [12,13]. Optimization techniques can incorporate design rules in terms of constraints and objectives that reflect the desired characteristics and performance of the design. Optimization techniques can also provide a way of evaluating design alternatives by comparing their objective values and ranked them according to their optimality. However, optimization techniques also have challenges, such as difficulty defining and quantifying the objective function, the trade-off between multiple conflicting objectives, and the computational cost of finding the optimal or near-optimal solution. Traditional approaches to building design include topology optimization, genetic algorithms, cellular automata, and generative and analogical design. These traditional methods [2] presuppose predefined rules for automated design and cannot learn and represent design codes and structural constraints outside predefined ones. So, this approach with hard-coded or restricted constraints cannot adequately represent design rules and design constraints.

Employing semantic models offers an alternative method for representing and assessing design intelligence. These models encompass the intended meaning and functionality of design solutions. They facilitate various analytical operations, including assessments of consistency, adherence to standards, or refinement through the application of logic-based techniques and deduction mechanisms [14–25]. Semantic models are also instrumental in measuring specific factors like interconnectivity, ease of access, or exposure by leveraging graph theories and network analyses. Semantic models confront several hurdles, such as constructing and upholding an exhaustive and uniform ontology, managing the uncertainty and diversity inherent in semantic details, and addressing the computational demands of reasoning algorithms. For example, graphs can model the topology and geometry of design layouts, as well as the attributes and constraints of each element [10].

One promising method to represent and evaluate design rules is using natural language, a natural and expressive medium for communication and reasoning. Language can capture the meaning and intention of the design solutions, as well as the context and preferences of the stakeholders. Language can also support various tasks and interactions, such as querying, explaining, or negotiating, to facilitate the design process. Language-based methods can potentially analyze and generate design layouts from textual descriptions or specifications [26].

**2.3 Generative Design with Rules**

Architectural layout design with rules can benefit from generative design methods that search a vast and varied design space, producing original and ideal solutions that meet multiple requirements and limitations. This section surveys the current research on architectural layout design with knowledge, concentrating on three main perspectives: rule-based methods, data-driven methods, and cross-modal methods.

Rule-based methods use predefined rules, constraints, and heuristics to generate and optimize layouts according to various criteria, such as functionality, aesthetics, and regulations [27,28]. These methods often rely on expert knowledge and domain-specific languages to encode the design rules and objectives. These methods can be divided into two sub-categories: shape grammar and optimization. Shape grammar methods use a formal language of shapes and rules to define and generate design alternatives [11]. Optimization methods use mathematical models and techniques to find optimal or near-optimal solutions to design problems with multiple objectives and constraints [29].

Data-driven methods use machine learning techniques to learn design rules and preferences from data, such as existing designs, user feedback, or contextual information, and generate layouts that reflect them [14]. These methods can be further divided into two sub-categories: generative methods and retrieval methods. Generative methods use probabilistic models, such as generative adversarial networks (GANs) [14–20], variational autoencoders (VAEs) [30,31], normalizing flows [32] , or graph neural networks (GNN) [33–36] to learn the latent distribution of design data and sample new designs from it. Retrieval methods use similarity measures, such as distance metrics, feature embeddings, or graph kernels, to find existing designs that match a given query or specification [35,37]. Knowledge-based design methods utilizing structured knowledge to inform and guide the design process have been a hot research topic in the AEC industry. Hu et al. [38] proposed Graph2Plan, which uses building boundaries as input to output floor plans and their associated layout graphs from the RPLAN database. Liu et al. [4] proposed embedding graph-constrained into the generative adversarial network for high-rise residential building design. Additionally, Jiang et al. [39], Huang and

Zheng [40], Nauata et al. [6], Tang et al. [41], and Wang et al. [42] employed GANs to learn from architectural design drawings and generate house layouts.

Cross-modal methods combine rule-based and data-driven methods to leverage the advantages of both approaches [43–48]. Cross-modal methods use multiple modalities of data, such as text, images, sketches, or speech, to represent and generate design layouts. These methods can leverage the complementary information from different modalities to enhance the design process and enable more natural and intuitive interactions. These methods aim to integrate the explicit knowledge and constraints of design rules with the implicit knowledge and creativity of design data. For example, Jain et al. proposed research to automatically render building floor plan images from textual descriptions [49]. Li et al. evaluate SD models' effectiveness in generating residential floor plans from text prompts [26]. Wu et al. and Jia et al. [50,51] use a conditional generative adversarial network to translate sketches into realistic floor plans.

In summary, existing methods for generating residential floor plans have limitations regarding the inputs to each model. Current models rely heavily on bubble diagrams, boundaries, and the addition of explicit constraints. The inputs to the bubble diagrams are mainly in graphical form, which is convenient for design professionals. However, for non-specialists, the lack of inputs such as shapes makes it difficult to have control when editing the generated residential planes. It, therefore, can be abstract and difficult to understand. Similarly, inputs in terms of room masks are more demanding as it is difficult for the designer to determine the location of the room, let alone the size and shape of the room, in the early stages of the design. Therefore, these models have limitations in terms of input flexibility.

**2.4 Research Gaps**

This review has identified several research gaps in the existing literature. First, although studies have refined the definition of design rules, they predominantly rely on rigid representations such as bubble charts, which limit the portrayal of design rules. Second, current models are confined to using visual graphs for both inputs and outputs, generally adhering to a single, fixed input method without offering flexible alternatives. The constraints of these representations necessitate considerable experience in the construction industry, as practitioners must prepare specific inputs like bubble diagrams in advance, creating a high entry barrier for non-specialists. Consequently, there is an urgent need for advanced models to generate designs adhering to essential design.

Sections 2.2 and 2.3 demonstrate the feasibility of using large-scale models with natural language to integrate essential domain knowledge and constraints for producing viable designs. However, research on architectural graphic design remains scant. Our study aims to address this gap, proposing a generative process that is more flexible and user-friendly for inexperienced designers. Design involves creating and communicating visual concepts that meet specific goals and constraints, often requiring both natural language and graphical inputs and outputs, such as sketches, diagrams, or layouts. For instance, a designer might need to generate a floor plan from a textual description of a client's needs or explain the rationale behind a design decision using natural language. Table 1 summarizes the main differences between existing methods and our proposed method for generating residential floor plans, comparing input modalities and output formats, and considering design rules or constraints. Compared to previous research, our method uses natural language as an input modality, making it more accessible and intuitive for non-expert users, and incorporates explicit and implicit design rules and constraints, such as room size, shape, adjacency, accessibility, and circulation, to ensure the designs' feasibility and functionality.

Table 1. Comparison of Existing Methods. (Position) Indicates that area or location implicit information is embedded in the bubble graph. Convolutional Message Passing (CMP), Boundary (B), Bubble Graph (BG), Layout (L), and Design Knowledge are shown in Table 2.

| Title | Method | Input | Extra Design knowledge | Design knowledge |
|---|---|---|---|---|
| ArchiGAN [52] | Pix2Pix | B | No | / |
| CoGAN [53] | Pix2Pix | L | No | / |
| House-GAN [5] | GAN+CMP | BG | Yes | (a), (c), (d)-(g) |
| House-GAN++ [6] | GAN+CMP | BG | Yes | (a), (c), (d)-(g) |
| Graph2plan: Learning floor plan generation from layout graphs [38] | GNN | B + BG | Yes | (a)-(c), (d)-(g) |
| Automated building layout generation using deep learning and graph algorithms [54] | U-net | B + BG | Yes | (a), (c), (d)-(g) |
| Residential floor plans: Multi-conditional automatic generation using diffusion models [55] | Diffusion model | Room mask (Position) | No | / |
| Architectural layout generation using a graph-constrained conditional Generative Adversarial Network (GAN) [3] | Pix2Pix+CMP | B+ BG | Yes | (a), (c), (d)-(g) |
| Our method | Large Diffusion Model | B or L + Natural language | Yes | (a)-(g) |

# 3. Problem Formulation

## 3.1 Design Requirements of Residential Layouts

In automated design generation, a critical challenge is ensuring that design outcomes adhere to both functional principles and practical architectural requirements. A literature review on automated building solution design, presented in Section 2.1, categorizes building design considerations into two main areas: design preferences and quality requirements. Design preferences tailor building solutions to the specific needs and desires of inhabitants, while quality requirements prioritize the building's functionality and occupant comfort. For abstract concepts such as architectural style, most current research relies on models to learn intrinsic architectural features. One of the objectives of this research is to propose a comprehensive mapping between graph structures and natural language based on the methods summarized in Section 2.4. Table 1 outlines design knowledge as characterized by different studies using graph structures, while Table 2 shows how this study incorporates seven traditional residential design criteria into the design generation process. Collectively, these elements form a comprehensive framework for evaluating and designing residential structures that meet both functional and aesthetic requirements while enhancing productivity.

Table 2. Residential Layout Design Rules

| Type | Requirements | Description or Example |
|---|---|---|
| Design preference requirements | (a) Number-related preferences | The desired number of rooms, bathrooms, balconies, etc. |
| | (b) Size-related preferences | The desired size for each room |
| | (c) Existence-related preferences | Availability of specific room types (e.g., master bedroom, guest room). |
| Design quality requirements | (d) Connectivity | All rooms should be well-connected. No isolated rooms |
| | (e) Circulation | Relationship between different rooms. Path complexity, length, etc. |
| | (f) Compactness | Efficient space utilization. Minimize empty or unusable spaces within the house. |
| | (g) Design norm | Ensuring living rooms are adjacent to kitchens for convenience or that bedrooms are positioned away from noisy areas. |

## 3.2 Rules Representation Approach

This section outlines a natural language-based approach to knowledge representation in residential graphic design. Traditional floor plan design primarily relies on image information, with knowledge representation confined to the construction of room masks or knowledge graphs. Room masks encode design rules via network operations that depend heavily on the specific model used, resulting in variable outcomes. Consequently, various techniques have emerged to integrate knowledge graph information into networks. For instance, methods such as graph neural networks and Conv-MPN within GAN models represent each room as a node, with connections between nodes to articulate design rules. Specific implementations, such as HouseGAN++ and Graph2plan, differ in their construction of knowledge graphs, which may include details on connectivity (via internal doors), location, proximity, and area. Traditional representations often lack flexibility and comprehensiveness due to model variations and limitations in the scope of expressible design rules. In contrast, natural language provides a more adaptable and comprehensive medium for expression (as shown in Fig. 2).

This article advocates using natural language to express architectural design rules while maintaining the structure of a knowledge graph to outline basic design requirements and ensure diversity. As detailed in Fig. 2, our approach segments the natural language descriptions into three parts aligned with the knowledge graph. The first segment specifies the number of various rooms, addressing points (a) and (c) in Table 2. The second describes the connectivity of the room, covering points (c) and (e)–(g). The third outlines room area ratios, meeting the requirements of point (b) as discussed in Section 4.2. Notably, this study does not differentiate between connectivity and proximity in the descriptions, anticipating that model training will intelligently resolve door placement in spatial layouts to define these aspects. This representation offers a cross-modal perspective of traditional knowledge graphs for graphic design; specifics on NLP are covered in Section 4.

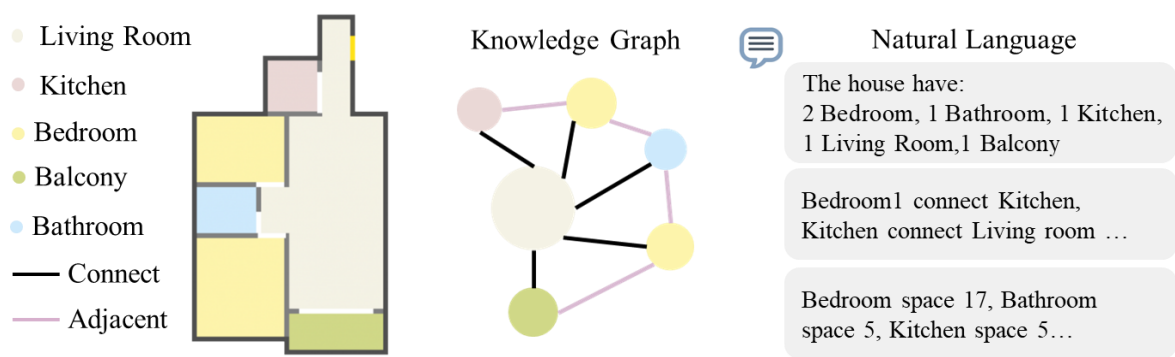

Fig. 2. Illustration of residential layouts' design rules and related nature languages

## 4. Methodology

This study proposes an innovative approach to automate the design of floor plans for residential layouts. As illustrated in Fig. 3, the methodology comprises two consecutive phases: (1) dataset preparation for residential layouts and (2) development of the residential layout generation model. The first phase involves preparing data by implementing a natural language-based representation of design rules. Specifically, the study integrates design rules into a knowledge graph and proposes a scheme to normalize the representation of the knowledge graph, thus enabling a standardized natural language-based representation of these rules. By transforming complex architectural knowledge into natural language, we offer a standardized and accessible way to represent design rules, reducing the need for extensive design expertise from users and lowering the design threshold.

In the second stage, we fine-tune the SD model for the residential layout generation task using a LoRA approach. Our method employs natural language as design constraints and incorporates ControlNet to generate controllable layouts through two distinct paths, each with explicit inputs and outputs:

(1) Text to Residential Layouts Generation: Preferences are provided via natural language text, and the preferred house type is supplied as an image. The text input includes general preferences, such as the number of rooms. The image input is the preferred floor plan. The model outputs a layout similar to the preferred type, adjusted based on the user-specified preferences, resulting in a new design that meets the user's needs.

(2) Boundary to Residential Layouts Generation: Targeted at professional designers, this path involves inputting detailed room connectivity relationships via natural language text and boundary images controlled by ControlNet. The text input specifies detailed design requirements, including specific room connectivity and spatial relationships. The image input consists of boundary constraints that define the limits of the design space. The model outputs a layout that strictly adheres to the boundary constraints and detailed specifications, providing a precise and controllable design solution.

Our approach offers flexibility by accepting multiple input types for learning objectives, enabling both professionals and non-professionals to directly express design requirements. The use of natural language for input lowers the design threshold, making the design process more accessible and user-friendly. By combining natural language and image inputs, we enhance the interpretability and flexibility of the design process, catering to a wider range of user needs.

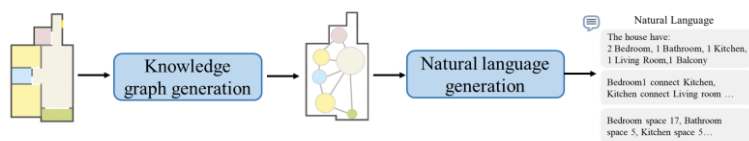
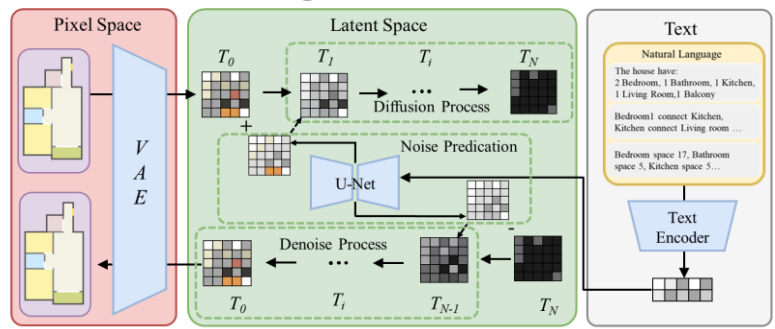
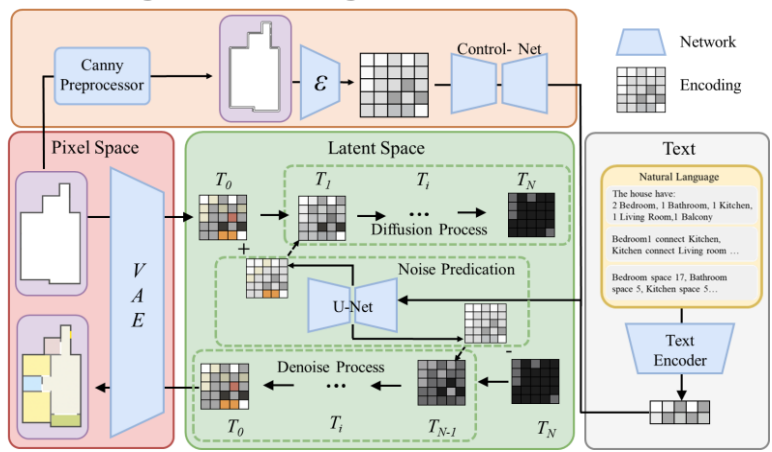

Fig. 3. The residential layouts' design method.

## 4.1 Dataset Preparation

This study utilized the RPLAN dataset, derived from authentic architectural drawings and containing over 80,000 unique building floor plans. The dataset was subjected to a standardized pre-processing procedure to ensure case diversity and enhance the visual-to-natural language representation of design rules, as illustrated in Fig. 4.

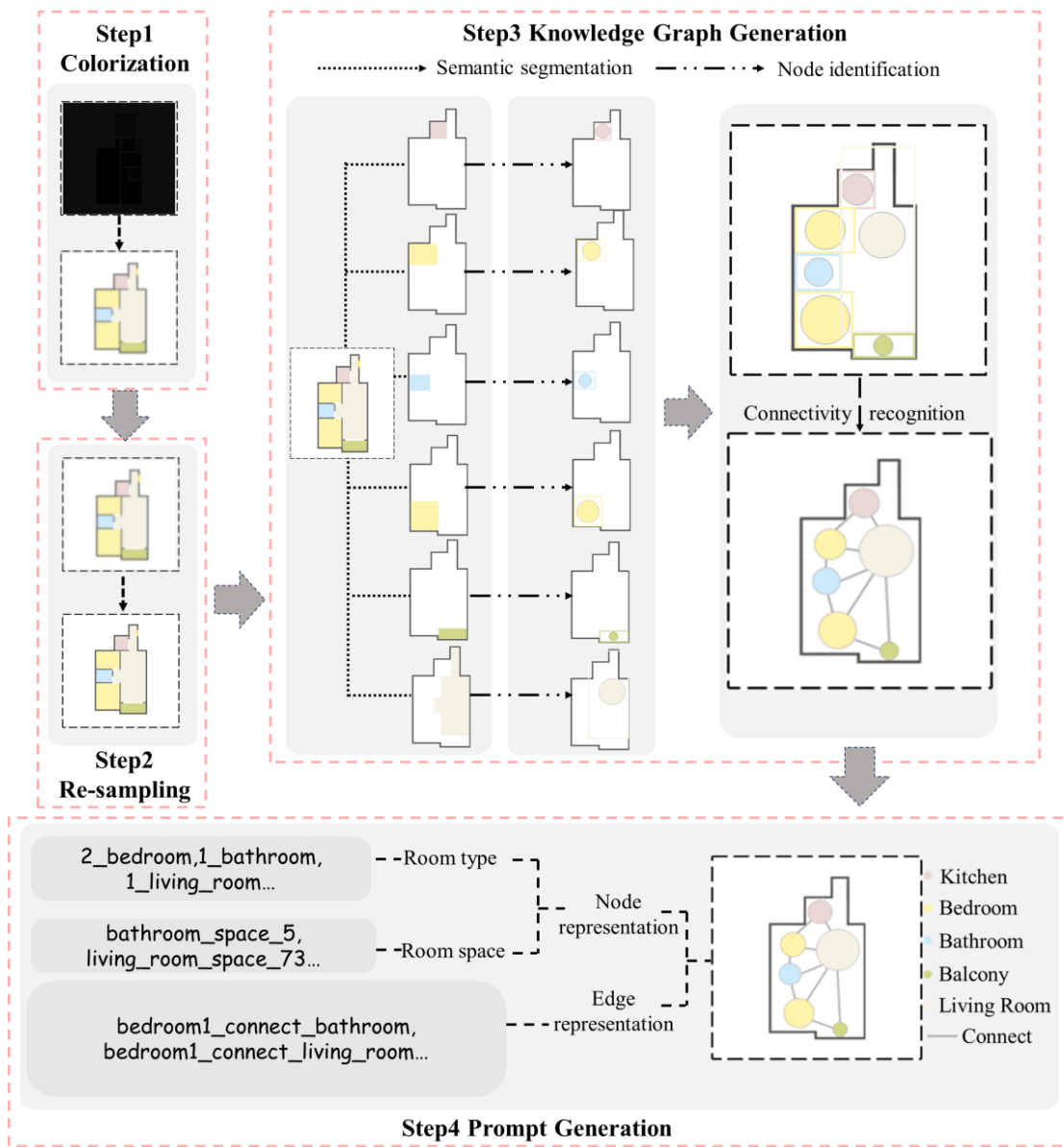

Fig. 4. Dataset preparation flowchart

The preprocessing encompassed four distinct steps: the initial two steps involved generating the plan layout, recoloring, and upsampling the RPLAN dataset to improve image quality, minimize the variety of room types, and reduce computational demands on the model. The third step included semantic segmentation and the creation of a knowledge graph that encapsulates the design rules of residential layouts. The final step involved translating this knowledge graph into a natural language representation, standardizing the expression of residential layouts' design rules. These steps collectively prepared the input data for cross-modal feature capture and facilitated the flexible generation of diverse residential layouts.

**4.1.1 Illustration of Dataset and Colorization**

Each sample in the RPLAN dataset is a 256x256 RGBA (Red, Green, Blue, Alpha) format image with four different channels. These images are not only labeled but also annotated through the four channels, highlighting their semantic information. Channel 1 is used to label external walls and front doors, and channel 2 is used to label the different spaces (room types) in the layout, as well as internal walls and doors. Channel 3 is used to show the number of repetitions for each room type in the layout, and channel alpha is used as a mask to distinguish between external and internal areas. In order to reduce the complexity of the design rules, the room types were simplified in this study, including the merging of rooms with similar functions, such as living room, dining room, and entrance, and the merging of uncommon room types as well, such as the merging of Wall-in with toilet. The 13 space types were reduced to 7, and different space types and different types of walls were recoloured. Table 3 shows the details of the merged space types and the specific recolouring data.

Table 3. Colour Coding And Channel 2 Values for Various Spatial Types in Dataset.

| Space label | Channel 2 value in the original dataset | R | G | B | A |
| --- | --- | --- | --- | --- | --- |
| Living room, Dining room, entrance | 0,4,10 | 244 | 242 | 229 | 255 |
| Master room, Child room, Study room, Second room, Guest room | 1,5,6,7,8 | 253 | 244 | 171 | 255 |
| Kitchen | 2 | 234 | 216 | 214 | 255 |
| Bathroom, Wall-in | 3,12 | 205 | 233 | 252 | 255 |
| Balcony | 9 | 208 | 216 | 135 | 255 |
| Storage | 11 | 249 | 222 | 189 | 255 |
| External area | 13 | 0 | 0 | 0 | 255 |
| Exterior wall | 14 | 79 | 79 | 79 | 255 |
| Front door | 15 | 255 | 225 | 25 | 255 |
| Interior wall | 16 | 128 | 128 | 128 | 255 |
| Interior door | 17 | 255 | 255 | 255 | 255 |

### 4.1.2 Resampling

Since the training and generation of large models require high image clarity, a pixel region-based image resampling strategy is used to enhance the clarity of the images. Specifically, the resolution of the original image is enhanced from 256 pixels to 512 pixels. By boosting the resolution, the detail and quality of the image can be enhanced, resulting in more accurate and detailed results for model training and generation. This method of increasing resolution can provide richer visual information for the large model, thus helping the model to better learn and understand the image content, which can effectively improve the performance and application effect of the model.

### 4.1.3 Knowledge Graph Generation

The design rules that can be articulated through traditional visual design methods vary, and this paper offers a comprehensive overview of these methods, categorized into three primary aspects as shown in Table 2: 1) expression of the number of room types; 2) constraints on the connectivity between rooms; and 3) constraints on the area and location of rooms. The automated extraction of the knowledge graph in this study involves splitting different rooms into distinct masks to complete the type segmentation within the node information. Each mask represents a node, with its mask colour indicating the room category. The second step calculates the boundary frames of each mask and uses the center coordinates of these frames as the node positions, thereby storing positional information. The radius of each node is determined by the length of its bounding box width. Finally, rooms are merged to determine whether there is an intersection between the bounding boxes; if so, they are deemed connected, thus completing the edge set identification.

### 4.1.4 Prompt Generation

This section explains the method for generating natural language cues from a knowledge graph representation of a floor plan. The description structure adopted in this study initially presents the room type, followed by the room area, and concludes with the connectivity relations between rooms, reflecting the structure of the knowledge graph. Initially, the room types and areas represent the node information in the knowledge graph. Subsequently, the connectivity relations between these nodes are detailed. This description is structured as follows:

(a) Room Types: We employ a generalized structure that combines a broad description of room types with a specific enumeration of the number of each room type. This approach imposes constraints that limit and define the total number of rooms by including an additional cue word about the total number of rooms. This starting point facilitates subsequent descriptions of room sizes and connectivity without reiterating the total room count.

(b) Room Sizes: We utilize a non-prescriptive, itemized description of individual room sizes. Rooms of the same type are differentiated by numbering (e.g., Bedroom_1 and

Bedroom_2 as listed in Table 4), which clarifies that there are multiple rooms of the same type and maintains consistency with the initial room type description. Room areas are linked to their respective identifiers with an underscore, thus connecting the area measurement directly to the specific room.

Table 4. Illustration of Prompts for Design Knowledge Representation

| Residential layout with Knowledge Graph | Prompt |
|---|---|
| 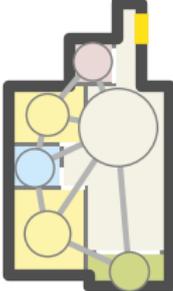 | The room has 2_bedroom, 1_bathroom, 1_living_room, 1_kitchen,1_balcony, bedroom1_space_17, bathroom_space_5, living_room_space_73, bedroom2_space_11, kitchen_space_5, balcony_space_7, bedroom_1 connect bathroom,bedroom_1 connect living_room,bedroom_1 connect balcony, bathroom connect living_room, bathroom connect bedroom_2,living_room connect bedroom_2,living_room connect kitchen, living_room connect balcony,bedroom_2 connect kitchen. |
| 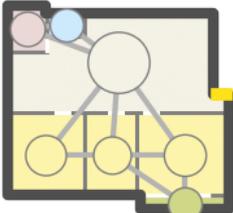 | The room has 2_bedroom,1_bathroom,1_living_room,1_kitchen,1_balcony,bedroom1_space_17,bathroom_space_5,living_room_space_73,bedroom2_space_11,kitchen_space_5,balcony_space_7, bedroom_1 connect bedroom_2,bedroom_1 connect balcony,bedroom_1 connect living_room,bedroom_2 connect bedroom_3,bedroom_2 connect balcony,bedroom_2 connect living_room,bedroom_3 connect living_room, kitchen connect bathroom, kitchen connect living_room, bathroom connect living_room. |
| 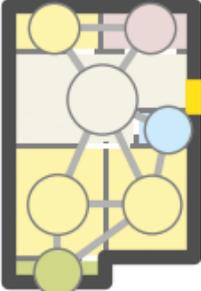 | The room has 3_bedroom,1_balcony,1_living_room,1_bathroom,1_kitchen,bedroom1_space_20,balcony_space_3,living_room_space_33,bedroom2_space_18,bathroom_space_3,kitchen_space_7,bedroom3_space_7, bedroom_1 connect balcony,bedroom_1 connect living_room,bedroom_1 connect bedroom_2,balcony connect bedroom_2,living_room connect bedroom_2,living_room connect bathroom, living_room connect kitchen, living_room connect bedroom_3,bedroom_2 connect bathroom, kitchen connect bedroom_3. |

(c) Room Connectivity: Having already detailed the type and size of each room, it is essential to describe the connectivity between rooms at the end of the sequence. This final step completes the description of the entire knowledge graph. Notably, while the room number (e.g.,

bedroom_1) may appear repeatedly, it always refers to the same node, ensuring consistency in feature description across the text.

## 4.2 Floor Plan Generation

This section delineates the methodologies employed to generate residential layouts using SD. The process bifurcates into two principal methodologies: Text-to-residential layouts generation and Boundary-to-residential layouts generation, as depicted in Fig. 5. Floor plans are initially represented in pixel space and are encoded into a latent space using a VAE. This latent representation is crucial for capturing the high-dimensional data in a more manageable form, which is then processed through the diffusion steps. The diffusion model is central to our approach to generating floor plans and consists of diffusion and denoise processes. The diffusion process involves progressively adding noise to the latent representations of the floor plans, moving them towards a purely noisy state over a series of time steps $T_0$ to $T_N$. At each step of the denoise process, a U-Net predicts the noise that has been added, which is then used to iteratively denoise the image, step by step, moving back from $T_N$ to $T_0$. After the denoising steps, the latent representations are converted back to pixel space using the decoder part of the VAE, resulting in the final floor plan that reflects both the textual and design requirements.

The generate method operates in two key conditioning pathways: textual descriptions and boundary constraints. Textual descriptions specifying room types, connections, and dimensions are first encoded using a text encoder. This encoded representation serves as a conditioning input to guide the diffusion process, ensuring the generated layouts align with the textual specifications. Boundary constraints are integrated using the ControlNet, which adjusts the generation process based on boundary images processed by a Canny preprocessor. Boundary images of floor plans are processed using a Canny edge detector to enhance boundary definitions, which are crucial for maintaining the structural integrity of the generated layouts. The processed boundary image is encoded into latent space via the encoded section ε in the VAE model, thus serving as an additional conditioning input for the control net. ControlNet works in conjunction with the U-Net architecture used in the diffusion process. It refines the denoised output from the diffusion steps by ensuring the floor plans adhere to the boundary conditions. This is achieved by adjusting the U-Net parameters dynamically based on the boundary inputs. The ControlNet fine-tunes the generated layout by minimizing the error between the predicted layout and the boundary constraints, ensuring the final output respects the specified boundaries. Combining U-Net's predictive capabilities with ControlNet's boundary adherence results in a floor plan that meets the aesthetic and functional requirements dictated by the natural language descriptions and the architectural boundaries.

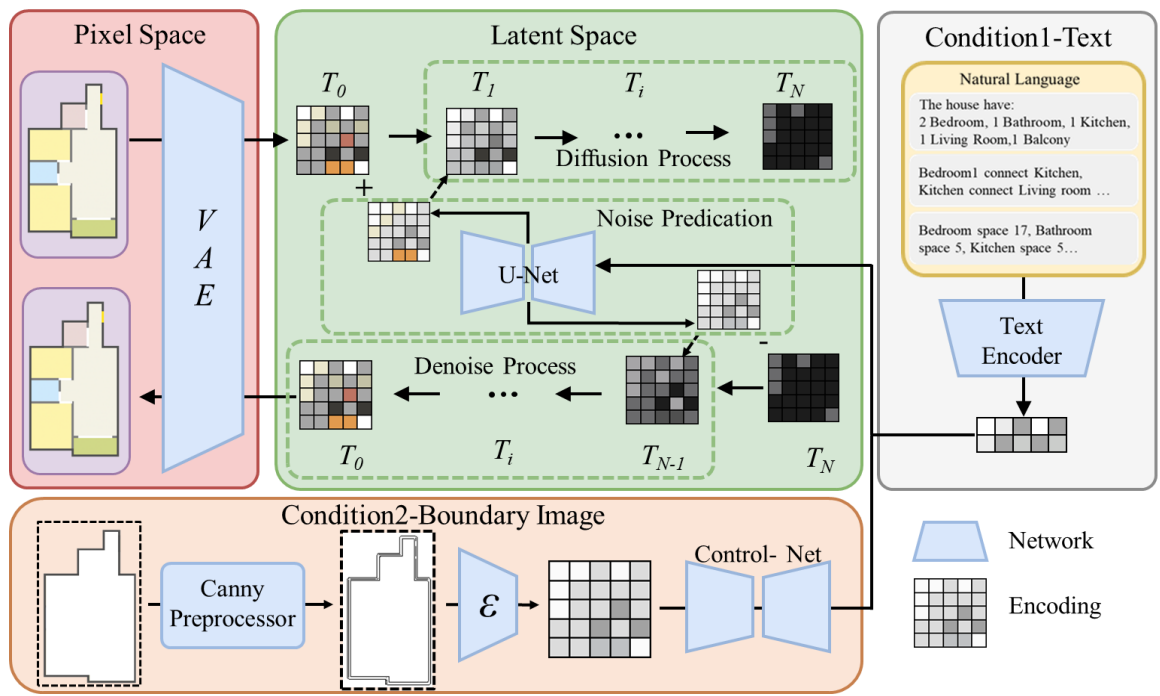

Fig. 5. Illustrations of two-path generation methods

### 4.2.1 Text to Residential Layouts Generation

This section elaborates on the first approach for generating floor plans, which uses textual descriptions alongside residential layout images as inputs to produce designs that adhere to specified layouts and conditions. The diffusion model, particularly the SD model, which is a variant of Latent Diffusion Models, excels in generating high-quality images from initial noise by iteratively refining noisy images into targeted designs. This model employs a reverse diffusion process that methodically transforms a purely noisy base into a clear, detailed image. The diffusion model functions through two primary processes: the forward diffusion process, which incrementally adds noise to the data as outlined in Eq. 1, and the reverse generation process, aimed at reconstructing data from noise, detailed in Eq. 2. Fig. 6 provides a schematic of the diffusion model's process.

$$q(X_t|X_{t-1}) = N(X_t; \sqrt{\alpha}X_{t-1}, (1-\alpha)I) \tag{1}$$

$$p_\theta(X_{t-1}|X_t) = N(X_{t-1}; \mu_\theta(X_t, t), \sigma_\theta^2(X_t, t)I) \tag{2}$$

Here, $X_t$ represents the data at time step $t$, and $\alpha$ is a parameter that controls the extent of noise addition. $\mu_\theta$ and $\sigma_\theta^2$ represent the mean and variance parameterized by network parameters, respectively.

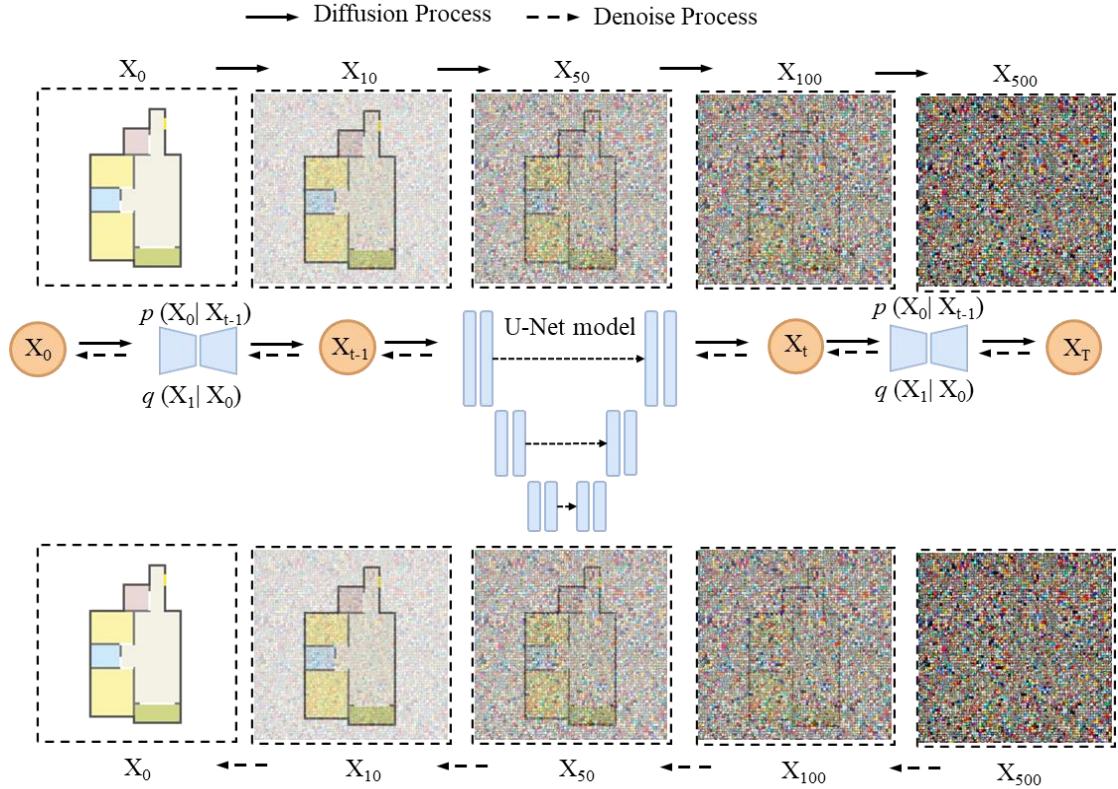

Fig. 6. Diffusion process

Diffusion models are conditional models that rely on prior information, typically in the form of text, images, or semantic maps. CLIP is employed to embed text or images into a latent vector T, shaping the final loss function, which also depends on the latent embeddings of the conditions. CLIP offers a robust method for encoding text comparably to images in a joint embedding space, essential for ensuring that generated layouts accurately reflect textual descriptions. In residential layouts generation, text embeddings from CLIP guide the diffusion process as conditional inputs, informing the U-Net of the desired attributes and relationships described in the text, such as room sizes and adjacencies. This conditioning guarantees that the generated layouts are not only realistic but also aligned with the specified textual inputs. CLIP employs a contrastive learning approach to map images and text into the same embedding space. This allows for the calculation of cosine similarity between image and text feature vectors, defined in Eq. 3.

$$sim(v,t) = \frac{v \cdot t}{\|v\|\|t\|} \qquad (3)$$

Where $v$ represents image embeddings, and $t$ represents text embeddings. By integrating the diffusion model with the advanced capabilities of CLIP for text encoding, the text-to-residential layouts generation pathway effectively translates complex textual descriptions into detailed and accurate floor plans. This method bridges the gap between textual concepts and visual representations, significantly enhancing the automation and precision of residential layout design.

**4.2.2 Boundary to Residential Layouts Generation**

This section elaborates on the second approach for generating floor plans, which uses textual descriptions alongside boundary images as inputs to produce designs that adhere to specified layouts and conditions. The ControlNet module is a fine-tuning technique applied to large models, particularly within the SD architecture, to enhance its ability to conform to specific boundary conditions while incorporating textual guidance. ControlNet adjusts the parameters of the U-Net within the SD model to align generated outputs with the provided boundary constraints. This adaptation ensures that the floor plans meet the aesthetic and functional requirements described in the textual input while strictly conforming to the specified architectural boundaries. Given the characteristics of the RPLAN dataset, which contains well-defined boundaries and distinct threshold variations, the Canny edge detection algorithm is used to preprocess boundary images. This preprocessing step enhances edge definition, facilitating ControlNet's ability to recognize and adhere to these boundaries during generation.

The preprocessed boundary images are then fed into the SD model, where ControlNet utilizes these enhanced edges to guide the diffusion process. The sharper and more defined the boundaries, the more effectively ControlNet can apply corrections to the U-Net parameters, ensuring that the generated layouts strictly conform to the boundary conditions. Fig. 7 illustrates the network schematic of the ControlNet module integrated within the U-Net architecture. ControlNet operates by dynamically adjusting the parameters of the U-Net layers based on input from boundary preprocessing, with these adjustments formulated mathematically in Eq. 4.

$$W' = W + \alpha \cdot ZeroConv(C(B)) \qquad (4)$$

Where $W$ is the original weight of the U-Net layers. $W'$ is the updated weight after applying ControlNet adjustments. $\alpha$ is the learning rate or scaling factor for updates. Zero convolution operation that adapts the control signals to the dimensions of the U-Net parameters. $C(B)$ is derived from the boundary information $B$, processed by the Canny edge detector. $B$ is a boundary image processed to enhance edge definition.

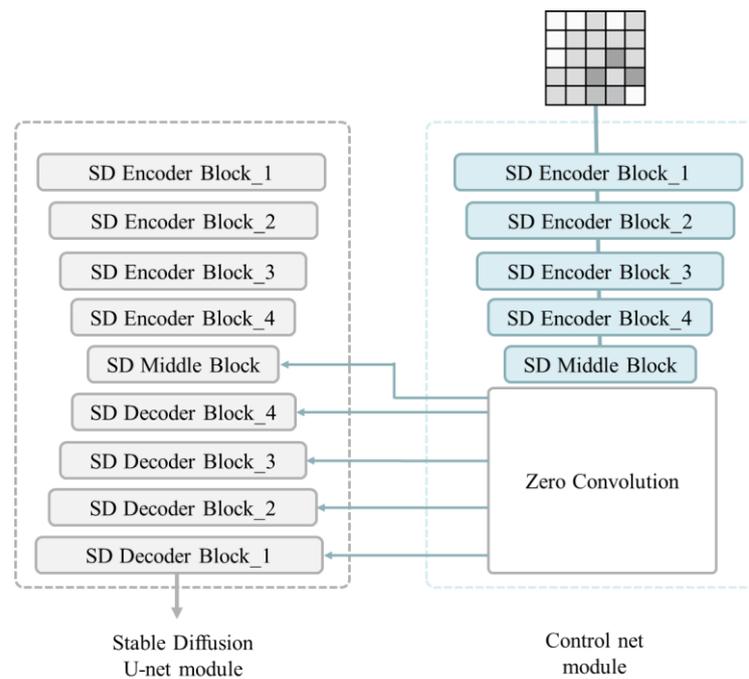

Fig. 7. Control net module

Fig. 8 illustrates the application of the ControlNet within the overall SD model framework. Each layer of the U-Net is fine-tuned in real-time to adapt to the minute details and constraints specified by the boundary image, ensuring that the final output retains the desired architectural features dictated by both the boundary image and the textual descriptions. By leveraging the ControlNet module within the SD model, enhanced by the Canny preprocessor, this pathway ensures that generated floor plans not only fulfill the aesthetic and functional requirements from the textual input but also adhere precisely to the physical boundary constraints. This method showcases the advanced capabilities of integrating fine-tuning techniques like ControlNet with edge detection algorithms to produce highly accurate and compliant architectural designs, paving the way for more precise and automated floor plan generation in architectural design tasks.

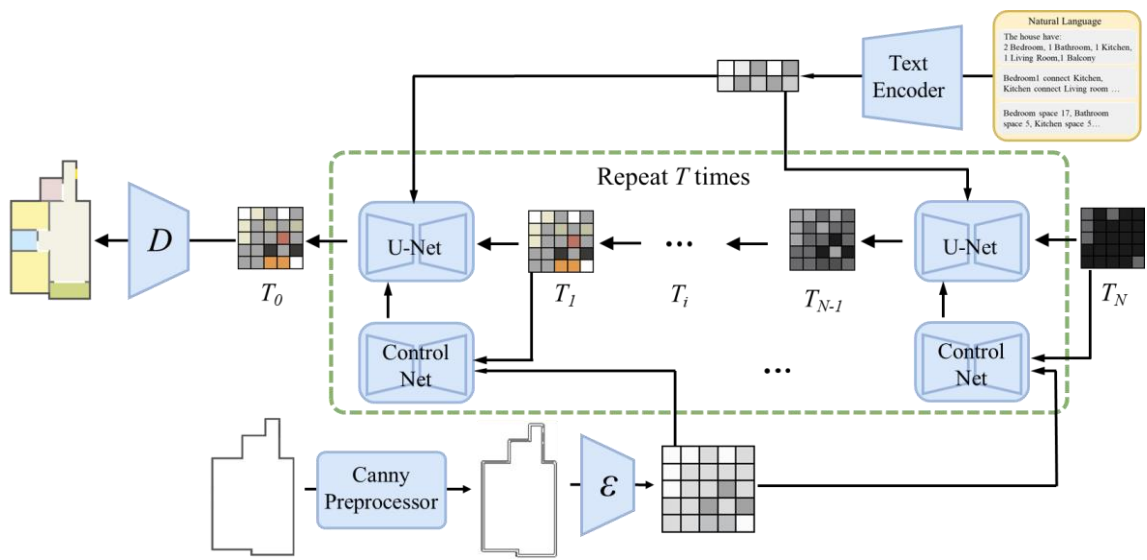

Fig. 8 Control net module Diffusion process

### 4.2.3 Model Training

However, existing diffusion models cannot generate realistic images of residential layouts. Therefore, we introduce a fine-tuning method based on LoRA to enable the diffusion model to learn the style characteristics of residential layout design rules. LoRA is an optimization method that reduces the parameter count and computational complexity of weight matrices in neural networks through low-rank decomposition. The basic idea is to decompose the pre-trained weight matrix into two low-rank matrices and introduce a scaling factor to control the update magnitude. Given an input x, the output y can be represented in Eq. 5.

$$y = x\left(W' + \frac{r}{r'}(A \times B)\right) + b \qquad (5)$$

Where *y* is the output, *x* is the input, *W'* is the pre-trained weight matrix, *A* and *B* are low-rank decomposition matrices, *r* is the scaling factor, *r'* is the rank of the low-rank matrices, and *b* is the bias term. The core idea of LoRA fine-tuning is to update the pre-trained weight matrix *W'* using the product of low-rank matrices *A* and *B*, with the update magnitude controlled by a scaling factor *r*. During fine-tuning, the pre-trained weights *W'* are frozen, and only the low-rank matrices *A* and *B* are trained. This setup allows the model to adapt to new data distributions while maintaining its original performance. This approach effectively reduces the number of parameters that must be optimized during training while preserving the model's performance.

A diagram of the LoRA module in action with u-net is shown in Fig. 9. Each conditioning step, whether it involves inputs from the image data or external conditioners like text descriptions, passes through LoRA-modified layers. These layers adjust their response dynamically based on the external conditioning inputs, which could be textual descriptions of the desired layout features or direct image-based modifications like boundary adjustments. Before and after the application of LoRA, the data undergoes normalization and activation, ensuring that the modifications made by LoRA are scaled appropriately and activated to enhance non-linear feature combinations beneficial for the generation process. Eq. 6 gives the loss function used for LoRA training, including a component that measures the difference between the model output and the target and a term that controls the magnitude of the update to prevent overfitting.

$$L_{LoRA} = \lambda_1 L_{task}(y, \hat{y}) + \lambda_2 L_{reg}(A, B) \qquad (6)$$

Where $L_{LoRA}$ is the total loss function for LoRA training, $L_{task}$ is the task-specific loss component (e.g., mean squared error for image generation) that measures the difference between the generated images $\hat{y}$ and the target images $y$, and $L_{reg}$ is the regularization term applied to the low-rank matrices *A* and *B* to prevent overfitting and control the magnitude of updates, typically through norms like the L2 norm. The hyperparameters $\lambda_1$ and $\lambda_2$ balance the contributions of the task-specific loss and the regularization term

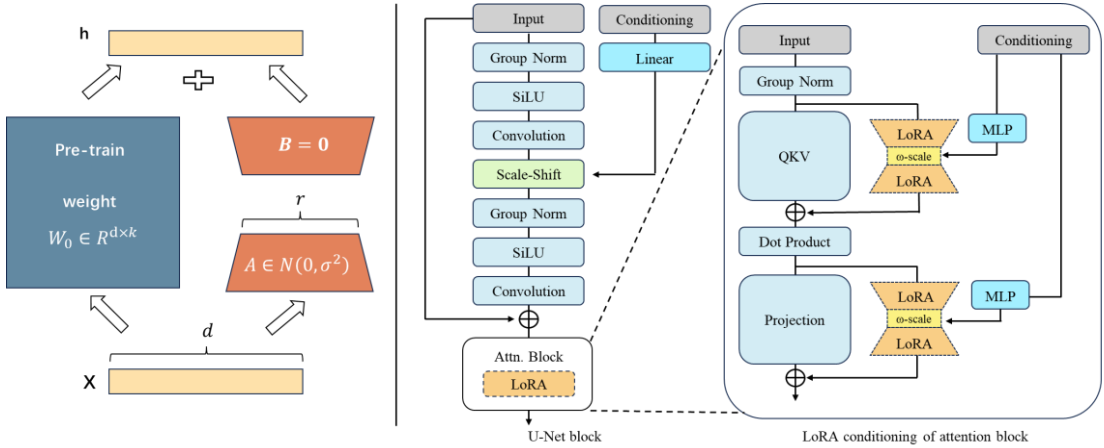

Fig. 9. LoRA module

## 5. Experimental Results and Discussions

In this study, pre-training for the SD model was done on a personal computer with an NVIDIA RTX A5000. The LoRA network dimension was set to 32, undergoing training across 10 epochs with a batch size of 14. Notably, we opted against employing gradient checkpointing to strike a balance between memory consumption and computational speed. A consistent learning rate of 1e$^{-4}$ was applied, with a slightly reduced rate of 1e$^{-5}$ for the text encoder, managed via a cosine scheduler with restarts to facilitate gradual reduction over time. The training utilized the AdamW8bit optimizer for enhanced processing speed while maintaining the training precision at FP16 to optimize computational efficiency.

To validate the reliability and diversity of the methodology, this study utilized four metrics to evaluate the generated results, and two baseline models were compared. Specifically, the

Fréchet Inception Distance (FID), Peak Signal-to-Noise Ratio (PSNR), Structural Similarity Index (SSIM), and Learned Perceptual Image Patch Similarity (LPIPS) were employed to gauge the model effects, thus avoiding the assessment bias of using a single indicator. The baseline models included the basic SD1.5 version model without fine-tuning and the pix2pix model. The SD1.5 version of the model is used to validate the effectiveness of fine-tuning in the proposed cross-modal generation method. Comparison with traditional visual generation methods We used both House diffusion[7] and Pix2pix[56] models to prove the effectiveness of the method.

## 5.1 Evaluation Metrics

To evaluate the effectiveness of the proposed method, this study employs four widely recognized metrics in image generation: FID, PSNR, SSIM, and LPIPS. These metrics are utilized to assess the quality of image generation:

- *FID*: FID measures the quality of generated images by calculating the difference between the distribution of feature vectors from real and generated images. It quantifies this by comparing statistical distributions, as expressed in Eq. 8.

$$\text{FID} = (\mu - \mu_w)^2 + T_r(\Sigma + \Sigma_w - 2(\Sigma\Sigma_w)^{\frac{1}{2}}) \qquad (8)$$

Here, $\mu$ and $\mu_w$ are the means of the respective feature vectors of the original and generated images, respectively; $\Sigma$ and $\Sigma_w$ are their covariance matrices; and $T_r(\cdot)$ denotes the matrix trace. A lower FID score suggests a higher similarity between the generated and real images, indicating superior image quality.

- *PSNR*: PSNR evaluates image fidelity by calculating the Mean Squared Error (MSE) between the original and reconstructed images, as shown in Eq. 9.

$$\text{PSNR} = 10 \cdot \log_{10}\left(\frac{MaxValue^2}{MSE}\right) \qquad (9)$$

Here, *MSE* is the mean squared error between the images, and *MaxValue* is the maximum pixel value, such as 255 for an 8-bit image. A higher PSNR value indicates a lesser disparity between the reconstructed and original images, thus denoting better image quality.

- *SSIM*: SSIM assesses image quality by analyzing luminance, contrast, and structural similarities between the original and reconstructed images, as detailed in Eq. 10.

$$\text{SSIM} = (x, y) = \left(\frac{2\mu_x \mu_y + c_1}{\mu_x^2 + \mu_y^2 + c_1}\right) \cdot \left(\frac{2\sigma_{xy} + c_2}{\sigma_x^2 + \sigma_y^2 + c_2}\right) \quad (10)$$

Here, $x$ and $y$ represent the original and reconstructed images, respectively. The terms $\mu$, $\sigma^2$, and $c$ denote the average pixel values, the variances of the images, and small constants that stabilize the division with a weak denominator, respectively. An SSIM value close to 1 suggests a high degree of similarity in terms of brightness, contrast, and structure, implying a high-quality reconstruction.

- *LPIPS*: LPIPS gauges image quality by computing differences in deep feature representations, using the VGG network to analyze subtle variances perceptible to the human eye. The core principle of LPIPS is to measure perceptual similarity by evaluating the distances between feature maps processed through the VGG network, as formulated in Eq.11.

$$\text{LPIPS}(x, y) = \sum_l \omega_l \cdot \|\emptyset_l(x) - \emptyset_l(y)\|^2 \quad (11)$$

Here, $x$ and $y$ are the image pairs being compared, $\emptyset_l(\cdot)$ represents the feature extraction function at layer $l$ of the VGG network, $\omega_l$ are the weights assigned to each layer, and $\|\cdot\|^2$ denotes the Euclidean distance. lower LPIPS score indicates higher perceptual similarity and usually superior image quality.

## 5.2 Training Results

### 5.2.1 Evaluation Metrics

In this study, we randomly selected 8,000 pairs of generated and real images to evaluate specific metrics. Table 5 presents the mean scores, while Fig. 10 illustrates the performance of the four metrics under different methodologies. FID scores suggest greater diversity in the model-generated results. Notably, the SD model trained with LoRA achieves scores significantly lower than the baseline model—up to 86.6% lower for both generation paths. Compared to the House diffusion method, the LoRA-trained SD model exhibits comparable diversity, with both methods achieving similar scores, indicating that the LoRA model effectively maintains diversity levels.

PSNR assesses image quality by calculating numerical differences at the pixel level, while SSIM measures image quality based on brightness, contrast, and structure. The SD model's results substantially surpass those of the Pix2Pix model in these metrics, which can be attributed to the diffusion model's superior architecture that produces higher-quality images compared to the GAN model. Compared to the House diffusion model, the SD model with LoRA also demonstrates superior image quality, achieving higher PSNR and SSIM scores, which reflect improved structural preservation and pixel-level accuracy.

Moreover, the improvements from LoRA training confirm that our method retains the inherent advantages of the SD model and ensures superior image quality compared to both the GAN model and the House diffusion approach. Additionally, LPIPS evaluates image similarity by analyzing the Euclidean distance within the feature space of a deep model; lower scores indicate higher similarity. The two-generation paths tested in this study achieved lower scores than the baseline and House diffusion, demonstrating effective capture of deep design rules. The LoRA model shows minimal fluctuation and high stability across all four metrics, reinforcing its robustness over both traditional GAN models and newer diffusion-based approaches like House diffusion.

Table 5. Results by Evaluation Metrics

| Methods | FID | PSNR | LPIPS | SSIM |
|---|---|---|---|---|
| Path1 with LoRA | 22.1 | 22.3 | 0.14 | 0.94 |
| Path1 without LoRA | 165.7 | 20.5 | 0.17 | 0.95 |
| Path2 with LoRA | 28.5 | 21.4 | 0.11 | 0.92 |
| Path2 without LoRA | 154.8 | 16.7 | 0.28 | 0.86 |
| Pix2pix | 142.6 | 4.8 | 0.65 | 0.46 |
| House diffusion | 32.6 | 13.6 | 0.45 | 0.78 |

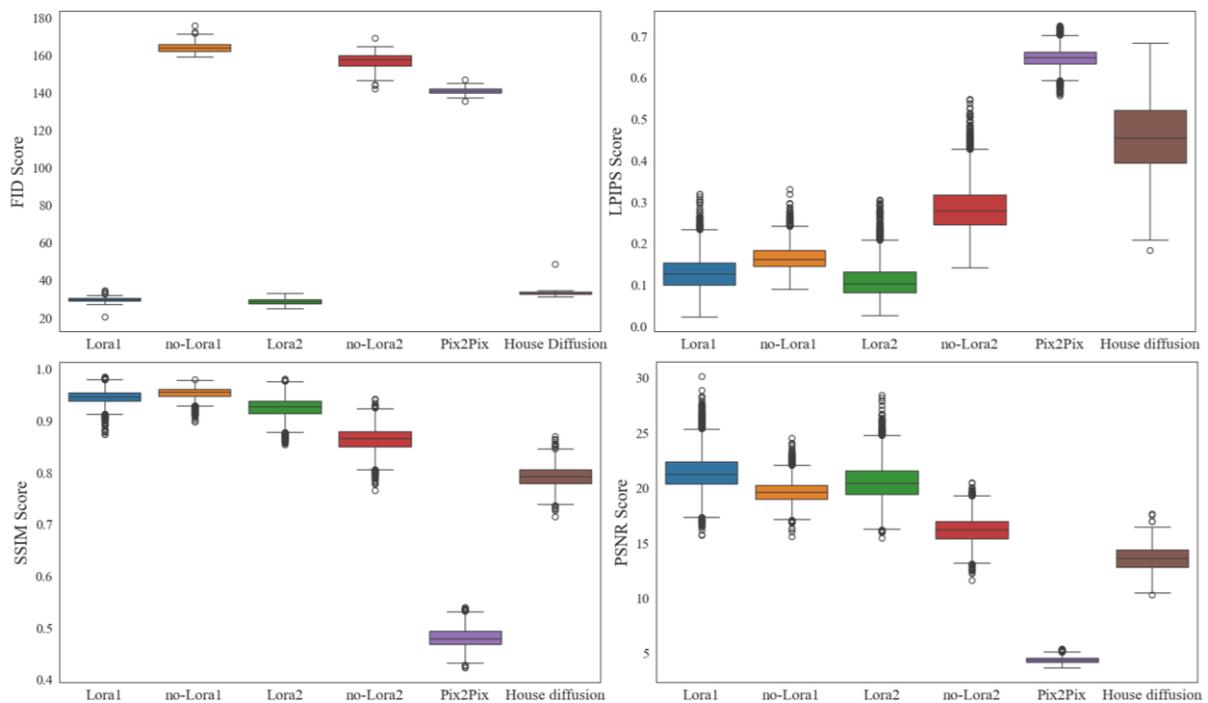

Fig. 10. Comparisons of Models' Performances in Evaluation Metrics

## 5.2.2 Method Comparisons

To validate the effectiveness of the proposed program, this study selected eight cases for comparison. Utilizing the RPLAN dataset, four cases were chosen from each of the two primary categories of three-bedroom and two-bedroom homes, with each case varying in the number of bathrooms and balconies. Specifically, the first case required one bathroom, the second case required one bathroom and one balcony, while the third and fourth cases required either two bathrooms or two balconies.

Figs. 11 and 12 illustrate the two cross-modal generation approaches proposed in this study using the LoRA model compared to the baseline SD model. These paths utilize distinct input image and text requirements as constraints for generative design. The first path employs a control net to manage building boundaries and text to define the functional relationships of rooms, aiming to generate solutions with consistent building boundaries. The second path leverages real images and text to jointly dictate the functional relationships, resulting in scenarios with similar boundaries and identical functional relationships. The images in Figs. 11 and 12 depict the input for each approach, and the corresponding text content is displayed in the Prompt column. It is evident that for Path 1, without LoRA, the base SD model fails to design layouts within the specified boundaries, while the LoRA model succeeds and strictly adheres to the prompt requirements. For Path 2, although the base SD model can produce a layout that somewhat resembles the boundary, the functional layout significantly deviates from the text description. In contrast, the LoRA model generates a building layout akin to the real image, with a graph structure representation that matches that of the real image. Additionally, the LoRA model demonstrates improvements in boundary overlap compared to the SD model, with a significantly higher overlap with the real image. The architectural design style of the LoRA model also aligns closely with that of the real image, indicating that the method of textual representation of design rules is feasible. Moreover, the fine-tuning of the SD model using the LoRA method effectively learns the design rules dictated by the text, enabling effective cross-modal building layout generation.

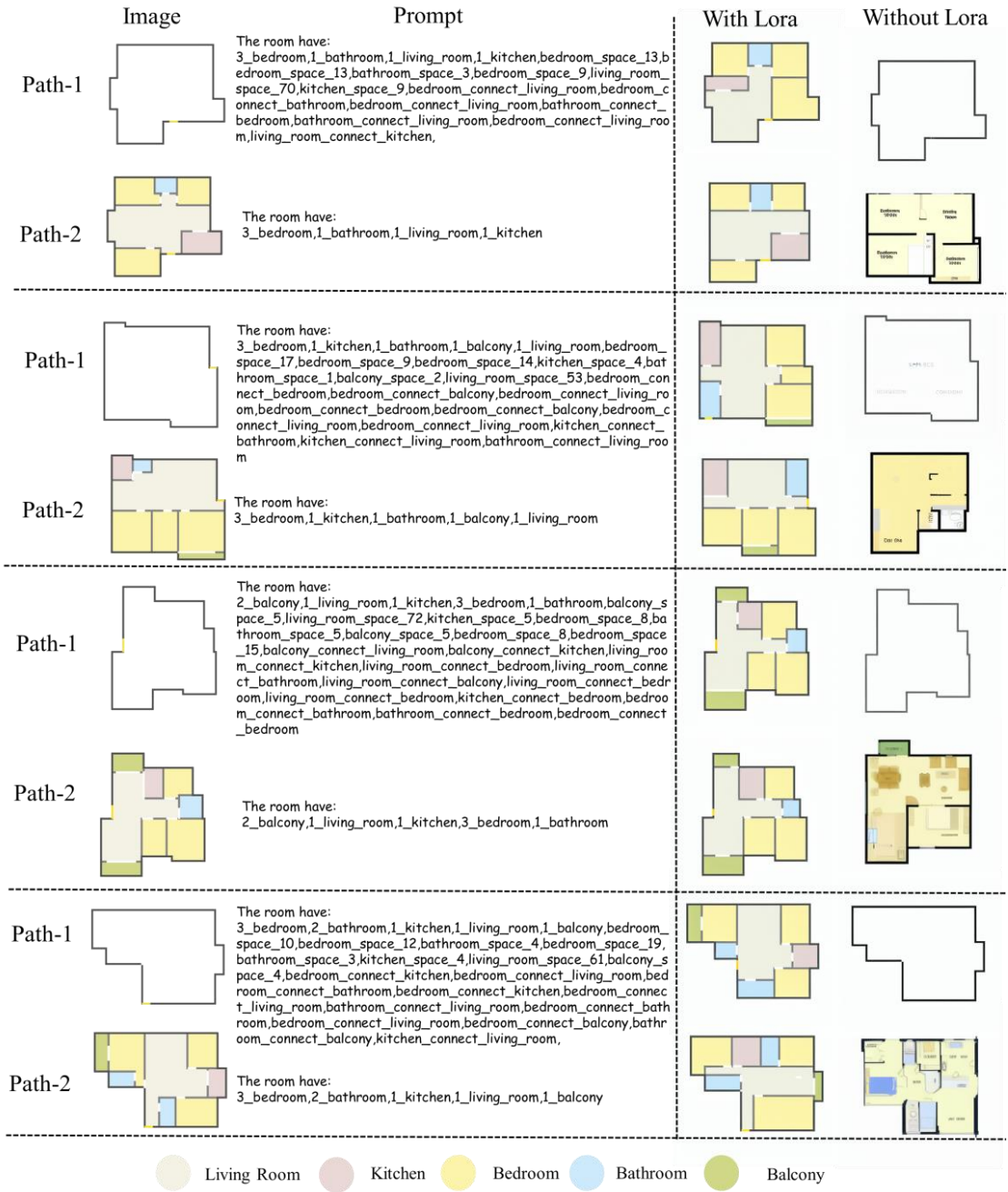

Fig. 11. Examples of three-bedroom generations through two-path approaches

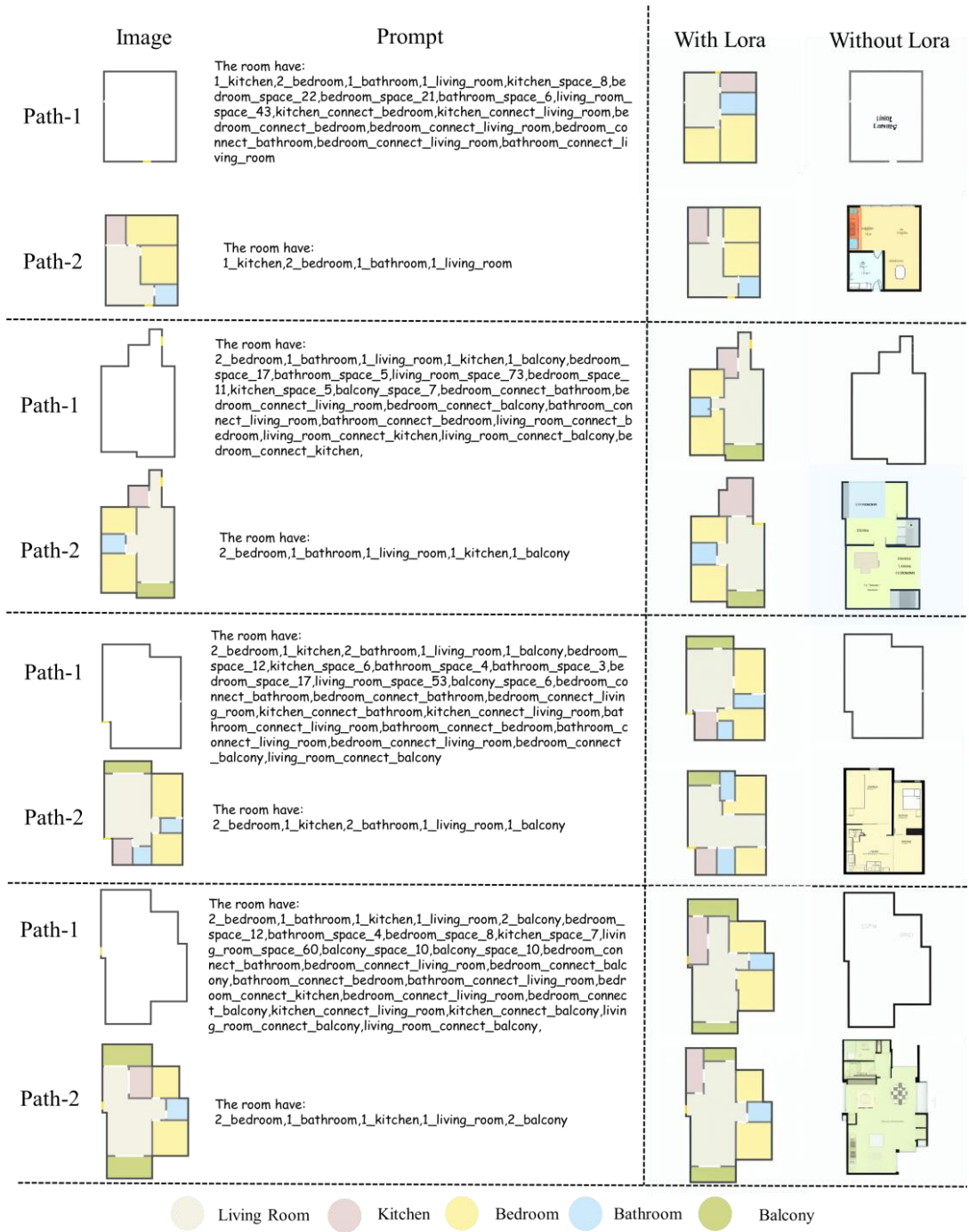

Fig. 12. Examples of two-bedroom generation through two-path approaches

Fig. 13 compares the results generated by the LoRA model with those from the Pix2Pix model and the House Diffusion model. While the boundary overlap between the Pix2Pix model and the real images is comparable to that of the LoRA model, the Pix2Pix model generates images of significantly lower quality. This difference stems from the inherent limitations of the Pix2Pix (GAN) model architecture, which makes it difficult to precisely control the number of each design element within the boundary, especially for detailed features such as doors and windows. Pix2Pix often fails to generate layouts with the correct number of rooms and appropriate connectivity, resulting in inaccuracies that do not meet design requirements. Additionally, the results from the Pix2Pix model tend to be more unstable, with a higher probability of stray lines appearing in the rooms, which further reduces the clarity and overall quality of the generated floor plans. Although the House Diffusion model generates higher-quality images than Pix2Pix, it also produces layouts with incorrect numbers of rooms or connectivity issues and cannot accurately constrain the boundaries, leading to significant deviations from the true boundary values. In contrast, the LoRA model exhibits superior accuracy, stability, and adherence to design constraints, resulting in cleaner, more consistent layouts.

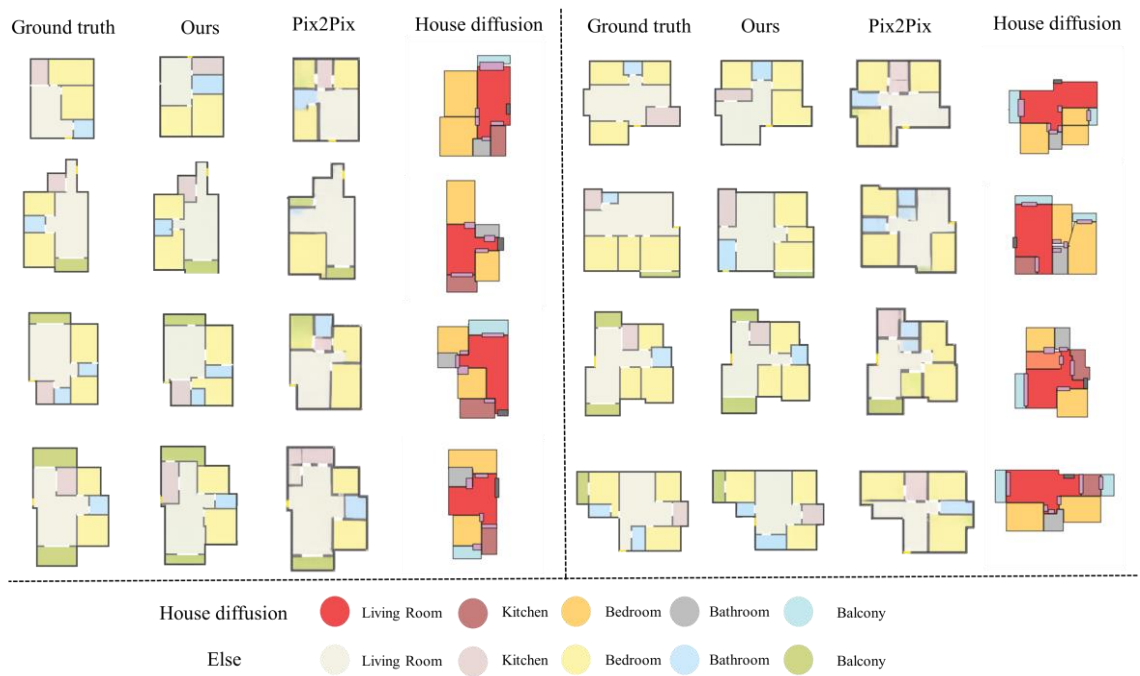

Fig. 13. Comparative analysis of generational outcomes between the suggested approaches and Pix2Pix and House diffusion.

### 5.3 Ablation Experiments

To assess the reliability of linguistic representations and the robustness of model-generated effects, this study explored the model's ability to reshape semantic information diversely under various input conditions. It also examined whether the model could adjust specific semantic information when provided. For Path One, the input design rule constraints comprised two elements: text and boundary images. The text outlined three components: room type, room connectivity, and room area, resulting in four design constraints that were tested sequentially. Initially, the model's controllability was evaluated, but some area information was missing. Four experimental groups were compared against the original results; the first group lacked a complete area description, while the second to fourth groups omitted details about the living room, balcony, and bathroom, respectively.

### 5.3.1 Experiments of Missing Area Descriptions

The chosen experimental case involved a complex layout from the RPLAN dataset featuring three rooms, two bathrooms, and two balconies, with results depicted in Fig. 14. Omitting area information effectively removed corresponding constraints from the graph nodes in the structure, as illustrated by the red text and nodes in Fig. 14. In the A experiment, lacking area constraints allowed the LoRA model to generate nearly identically sized bedrooms more effortlessly. In the B experiment, which is generally larger than the living room, the LoRA model successfully generated layouts without specific constraints of the living room area. The C and D experiments revealed that the model generated balconies and bedrooms of almost identical sizes, indicating its inability to control area proportions under incomplete constraints. Furthermore, the last column in Fig. 14 shows model generation failures due to the inherent instability of the large model, resulting in outcomes that defy design constraints, such as incorrect room connectivity.

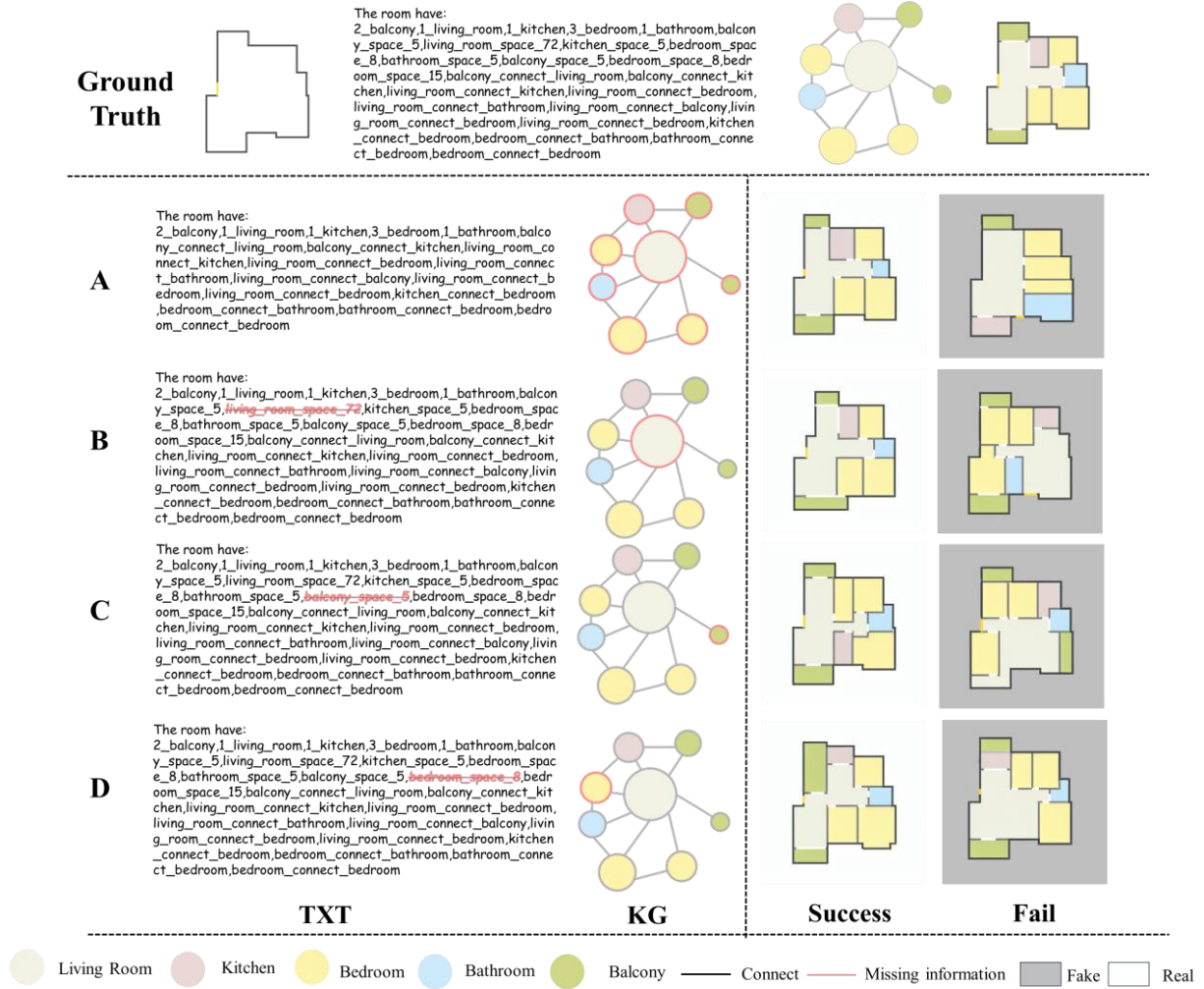

Fig. 14. Generation case of missing room area information

### 5.3.2 Experiments of Missing Room Connections

Fig.15 investigates missing room connection information. The A group in Fig. 15 lacked all connection details, with subsequent groups missing specific connections involving the living room and balcony, the living room and bedroom, or both. Despite missing complete connection data, the LoRA model managed to produce building layouts that met broader social requirements. However, the absence of specific constraints led to uncontrolled room connections, as seen in the first group's results in Fig. 15. The remaining three sets also reflected this, where the LoRA model freely generated the missing information, though not always aligning with the original design. For instance, balconies in the B and D groups are connected to the kitchen instead of the living room, and the balcony in the C group is connected to the bedroom.

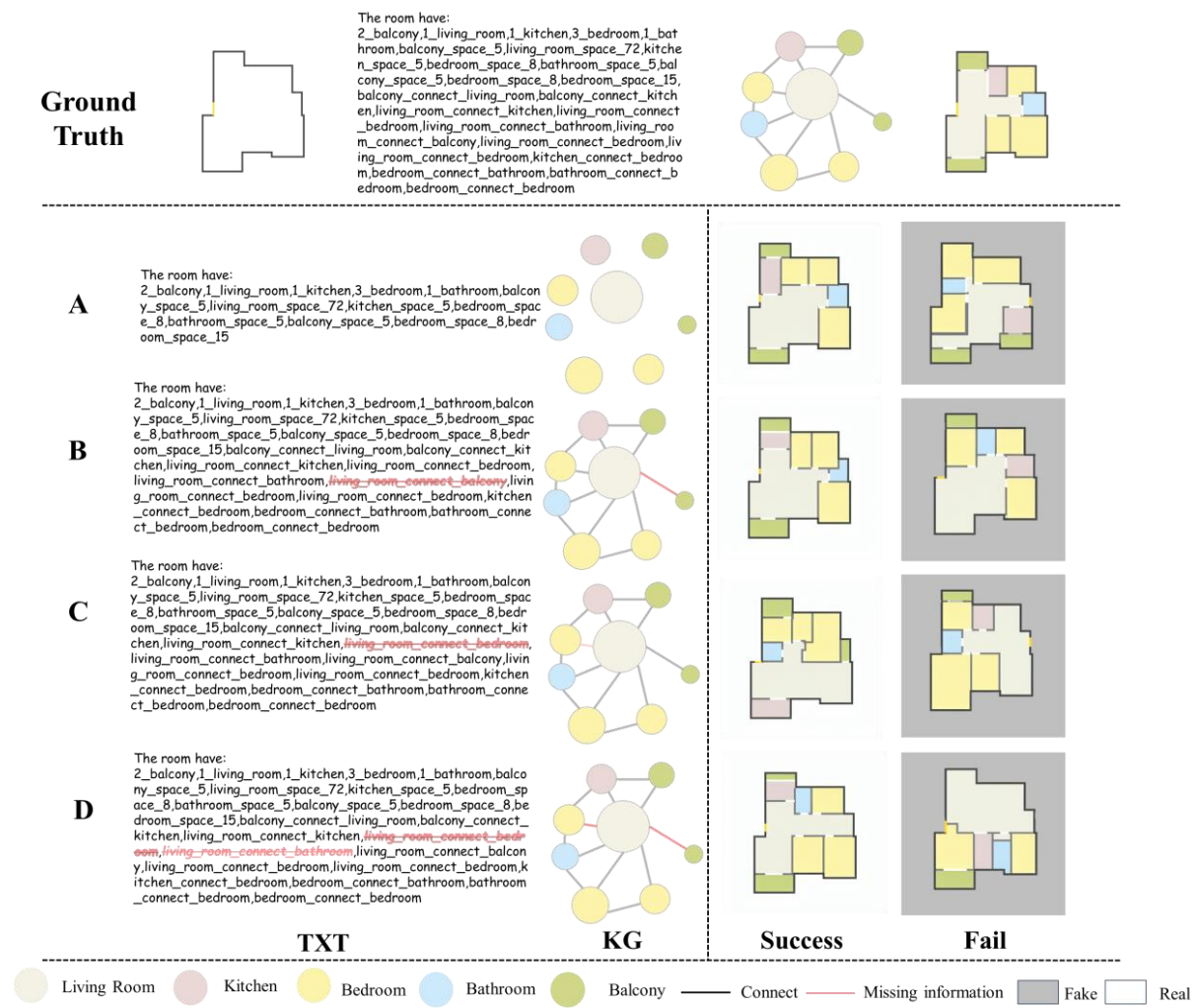

Fig. 15. Generation results of missing room connecting information

### 5.3.3 Experiments of Missing Area Descriptions and Room Connections

Fig. 16 displays generation results with both room connection and room area information missing—only room type, and number constraints remained. Following the same pattern, the A group lacked all the information, while the remaining groups lacked constraints on living rooms, balconies, and bedrooms, respectively. Unlike prior experiments with distinct area and connection constraints, missing nodes were represented by red dashed circles in the graph structure to indicate their randomness. The model generated scenarios that matched the number and type specified in the text, demonstrating its ability to adapt to user needs rather than merely replicating the original image, a capability not present in traditional GAN models. This indicates the model's potential to combine text and boundaries to create diverse new scenarios, albeit with some instability.

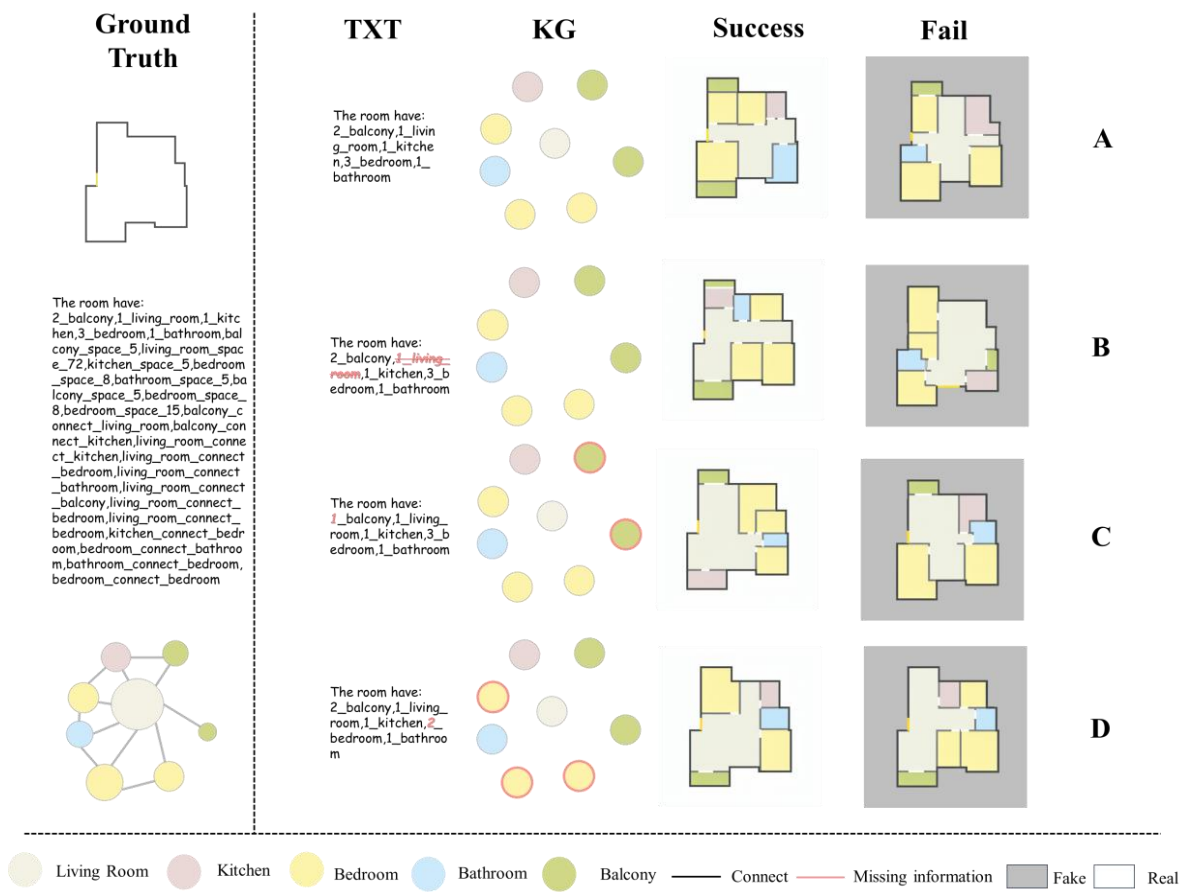

Fig. 16. Generation results on the number and type of rooms only

In conclusion, experimental studies have shown that missing semantic information allows for more diverse model generation. Models can fully exploit the creativity of large models within the constraints of remaining needs, providing users with more diverse and high-quality new scenarios.

The final study, shown in Fig. 17, was conducted on image variables using four different building boundaries with the same textual descriptions. The real images for these boundaries differed from the textual descriptions, while the generation results aligned with the textual descriptions for the same boundaries. This demonstrates the LoRA model's ability to adapt generation across different boundaries based on textual constraints, confirming the model's capacity for generating diversity.

### 5.3.4 Experiments of Different Room Boundaries and Types

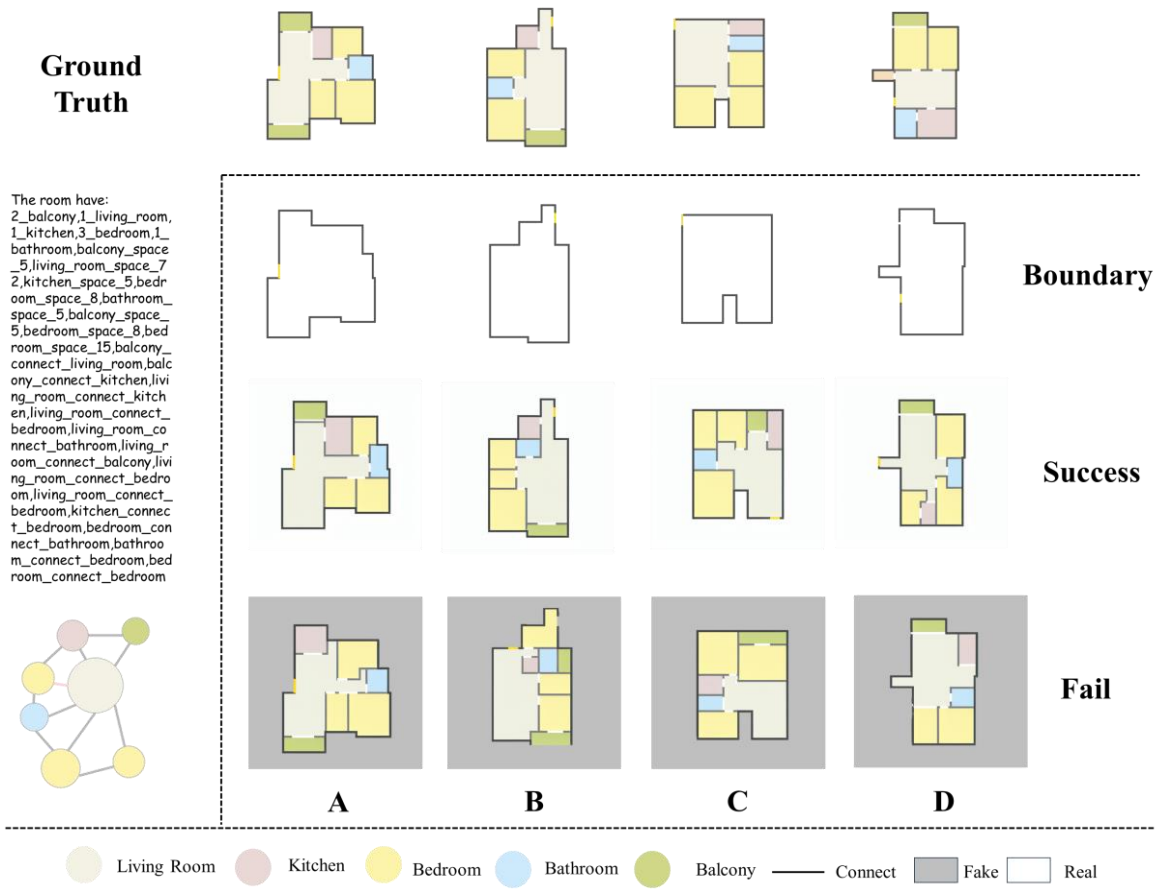

Fig. 17. Generation results for different room boundaries

### 5.3.5 Experiments of Missing Room Counts

Path 2 utilizes the actual layout as a reference for the functional relationships between rooms and employs the number of rooms as a constraint to generate new building designs. In this path, the only information that may be missing is the node information, as illustrated in Fig. 18. Similar to Path 1, this series of experiments lacks comprehensive data for the living room, balcony, and bedroom sequentially. When all information is absent, the model autonomously supplements the missing room node data from the original image to create varied building layouts. Given that the living room is an essential component of house design, the model automatically fills in any missing details for this room. Although balconies and bedrooms frequently change in architectural designs, the model produces layouts that adhere to the textual constraints specified. Notably, the failure cases in this path mirror those in Path 1, predominantly involving mismatches in-room connectivity. Both paths demonstrate that models can generate designs with controlled diversity, adhering to the input constraints.

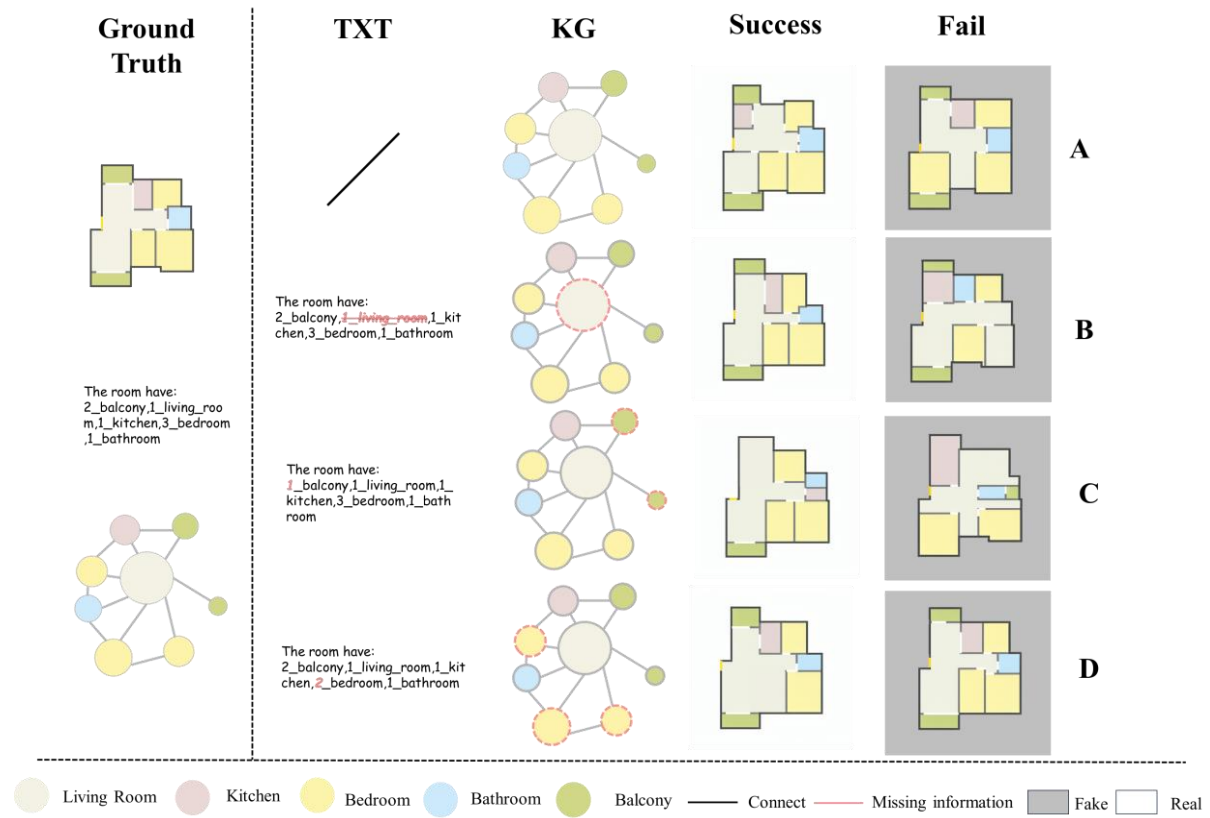

Fig. 18. Generation results for different room types

### 5.4 Experiment Discussions

This study proposes an approach for building layout design using a LoRA fine-tuned SD generation model with the linguistic representation of design rules for optimization. Compared with traditional deep learning frameworks such as GAN, this method significantly improves the flexibility and user-friendliness of the design process. By flexibly modifying the input constraints, users can adjust the design parameters according to their specific needs, thus enabling the design of complex building layouts without the need for deep expertise. In addition, the method also utilizes NLP technology to make the design instructions more intuitive and easier to understand, which greatly reduces the technical threshold of the design process.

However, current knowledge representations are still insufficient in terms of area and connectivity control, and there are still limitations to the generation of complex building layouts. Although initial progress has been made, there is still much room for improvement in the degree of matching and accuracy of design rules. For example, when dealing with large and complex spaces, the current model is often unable to accurately control the proportion and layout of each functional area, resulting in the deviation of the final design result from the user's expectation. In addition, further research and improvements are needed regarding the connectivity within the building, such as optimizing access layouts and spatial flow lines to ensure that the final design is aesthetically pleasing and practical. Meanwhile, due to the limitation of CLIP technology, the length of the cue word is restricted, and the problem of excessively long cue words may exist when facing uncommon house types with four or more complex apartments.

For residential buildings, our method simplifies the conversion process from conventional to intelligently generated layouts. SD models fine-tuned with LoRA not only support diverse design requirements but also respect the individual preferences of designers, allowing them to create residential spaces that match users' habits and cultural contexts while maintaining design quality. In this way, our approach not only improves the efficiency and quality of design but also expands the possibilities of architectural design, making it more personalized and user-friendly.

## 6. Conclusion and Future Work

There are several research studies on the varied and effective design of residential layouts. However, existing model architectures require a single input and lack flexibility in expressing design rules. To address these challenges, this study proposes an automated and controllable residential layout design process using the SD model with the functionality of ControlNet. This foundation facilitated the development of a cross-modal generation model, employing the LoRA technique for training to enable more intuitive user interactions through natural language inputs. We explored two distinct design paths: the first leveraged boundary constraints and specific room requirements, while the second adapted to user preferences' layout and variable room counts, demonstrating our model's capacity to understand and generate layouts that reflect both explicit instructions and inferred room connectivity. The evaluation phase confirmed the model's effectiveness and adaptability, showcasing its potential to change traditional residential design practices by accommodating a wider array of user needs and preferences. Based on this study, the conclusions are summarised as follows:

1. This study summarizes the design rules characterized by traditional deep model architectures and provides a comprehensive representation of the design rules using natural languages. This study defines the mapping scheme between knowledge graphs and natural languages. This mapping enables the effective use of natural language to describe design requirements, facilitating the control and refinement of design processes. By integrating knowledge graphs and NLP, the approach allows for a more intuitive and accessible representation of complex design rules, enhancing the ability to communicate and implement design specifications within architectural projects.

2. The research presents a cross-modal generation approach that enhances the training of large SD models with the LoRA technique. This algorithm helps the SD model to incorporate design principles efficiently, providing a useful method for implementing large models within the construction sector. Additionally, the refined SD model produces images of superior quality compared to conventional pixel-based generating techniques such as Pix2Pix and House diffusion methods in the metrics of FID, PSNR, SSIM, and LPIPS.

3. This study validates the adaptability and flexibility of the generative model under conditions of missing semantic information through ablation experiments. Experiments demonstrate that the model can leverage remaining information to generate diverse new scenarios that meet design needs, even when specific semantic information about room areas or connections is incomplete. Additionally, when facing missing room node data, the model autonomously supplements these data to produce varied but textually constrained architectural layouts. Although there were instances where the model produced results inconsistent with the original design, these experiments show that by flexibly adjusting input constraints, the LoRA fine-tuned SD model can fully exploit the creativity of large models while adhering to user requirements. This method not only enhances the flexibility and user-friendliness of the model but also improves the efficiency and quality of design, making architectural design more personalized and responsive to user habits.

Regarding future research, while this research focuses primarily on residential layouts, the data-driven derivative design framework focuses on early-stage spatial design considerations, making it applicable and relevant to a wide variety of housing projects, regardless of their construction methods. Considering aspects such as more frequent human-computer interaction design and more natural design adjustments, improving the linguistic representation of design rules to better incorporate additional design constraints like sustainability and energy efficiency are important future research directions towards more efficient energy use and sustainability in the AEC field. Additionally, extending the proposed method by integrating BIM systems and rich building information to further support the intelligent design process has also been identified as an important research topic.

## Acknowledgments

The authors would like to express their gratitude for the financial support provided by the Chongqing Technology Innovation and Application Development Special Key Project (Grant No. CSTB2022TIAD-KPX0136) and the National Natural Science Foundation of China (Grant No. 52130801, 52308141).